\documentclass[runningheads]{llncs}
\usepackage{graphicx}
\usepackage{booktabs}
\usepackage[ruled,vlined]{algorithm2e} 

\usepackage{algorithmic} 

\usepackage{verbatim}
\usepackage{microtype}
\usepackage{subfigure}
\usepackage{amsmath} 
\usepackage{hyperref}

\usepackage{amsthm}

\spnewtheorem{assumption}{Assumption}{\bfseries}{\itshape}

\usepackage{color,xcolor}
\usepackage{amsmath}\allowdisplaybreaks
\usepackage{amsfonts}

\begin{document}

\title{Conservative Contextual Combinatorial Cascading Bandit}
%
%
\author{Kun Wang\inst{1} \and
Canzhe Zhao\inst{1} \and
Shuai Li\inst{1} \and
Shuo Shao\inst{1}}

\authorrunning{Kun Wang et al.}

%
\institute{Shanghai Jiaotong University, Shanghai, China}

%
\maketitle              
\begin{abstract}
Conservative mechanism is a desirable property in decision-making problems which balance the tradeoff between the exploration and exploitation. We propose the novel \emph{conservative contextual combinatorial cascading bandit ($C^4$-bandit)}, a cascading online learning game which incorporates the conservative mechanism. At each time step, the learning agent is given some contexts and has to recommend a list of items but not worse than the base strategy and then observes the reward by some stopping rules. We design the $C^4$-UCB algorithm to solve the problem and prove its n-step upper regret bound for two situations: known baseline reward and unknown baseline reward. The regret in both situations can be decomposed into two terms: (a) the upper bound for the general contextual combinatorial cascading bandit; and (b) a constant term for the regret from the conservative mechanism. We also improve the bound of the conservative contextual combinatorial bandit as a by-product. Experiments on synthetic data demonstrate its advantages and validate our theoretical analysis.

\end{abstract}
\section{Introduction}
There are many problems in real world which can be formulated as decision-making problems under uncertainty. In this situation, multi-armed bandit (MAB) is extensively studied and used for it. It can solve the famous exploration-exploitation dilemma, which is common in many situations. The basic model of MAB is formulated as an online learning problem: each arm is associated with a noisy reward. At each time step $t$, the learning agent should choose one of $K$ arms to get the maximum cumulative rewards. We use the regret (the difference of cumulative rewards between the optimal strategy and the algorithm) to identify the algorithm's performance.

Recently, there are many variants of multi-armed bandit such as contextual bandit, combinatorial bandit, conservative bandit, cascading bandit and gaussian proccess bandit. These variants are the bandit problems applicable for some specific situations in order to have higher accuracy and stability. Especially, the cascading bandit and the conservative bandit start to attract more attention in recent years. The cascading bandit is an online learning process. At each time step $t$, the learning agent receives  context vectors for some arms. The expected weight of every arm is an unknown parameter $u$. To get the maximum cumulative rewards, the agent chooses a list of arms and receives the stochastic weights of every arm checked under some stopping rules. The core of this problem is the estimation of every arm's expected weight $u$, which incurs subsequent regret. It has many direct applications in recommender system and search engine.

The contextual bandit also shows its advantages in understanding the nature of the bandit problem. Contextual bandit is a sequential decision-making problem where in each time step, the learning agent chooses an arm and gets a stochastic reward in return, expected value of which is an unknown linear function of the context and the parameter $\theta$. Thus the core of this setting is to estimate $\theta$. Because it includes the user and the arm information and has a high-dimension feature, it's extensively used in the personalized recommendation. The setting has been proposed long time ago, but the proof of the upper regret bound consumes for a long time. Eventually, Abbasi et al.\cite{abbasi2011improved} successfully prove its $O(\sqrt{n})$ upper bound.

Although many learning algorithms are superior to solve the bandit problem, most of them don't guarantee the performance of initial exploration phases. This is a big barrier in many fields such as online marketing, health science, finance and robotics. Therefore, the learning algorithm with safety guarantee can tremendously increase the applicability in solving decision-making problem. The conservative bandit becomes more and more attractable in such a safety-concerning world. In this setting, we already have a base strategy and wish to design an algorithm better than it. In the basic bandit, the setting doesn't take the conservative mechanism into consideration. But in 2016, Wu et al.\cite{wu2016conservative} proposed the conservative bandit to incorporate the safety and proved the high-probability upper regret bound for the problem in both stochastic and adversarial settings. In conservative bandit, the expected reward of the baseline strategy is random, so the safety guarantee holds with high-probability. However, this model merely illustrates the simple basic bandit with conservative mechanism and there remains a lot of sophisticated model to fit into such as contextual, combinatorial and cascading bandit, etc. 

This paper incorporates all of the above feedback so that it can apply to real scenarios more appropriately. Aimed at safety, we first formulate this setting named as the conservative contextual combinatorial cascading bandit ($C^4$-bandit) in Section 3. In Section 4, we propose an algorithm based on the \emph{Optimism in the Face of Uncertainty (OFU)} to solve it, named $C^4$-UCB. In Section 5, we prove the upper bound for the $C^4$-UCB in two situations which can be decomposed into two terms. The first term is an upper bound for the $C^3$-UCB, which doesn't have the conservative mechanism, growing with the time $T$. The second term is a constant (not growing with the time $T$) and accounts for the loss of being conservative. Finally, in Section 6, we show our experiment results of $C^4$-UCB, and validate our theoretical analysis.

This paper has a few advantages and creativity. First, it incorporates as many settings as possible including context, combinatorial and cascading. Second, it's combined with conservative mechanism in the cascading bandit and get the exact from of the regret for the first time so that it can ensure the safety of the exploratory phase in the recommender system. Last but not least, it first make the regret from the conservative constraint bounded by a constant term in the combinatorial bandit that Zhang et al.\cite{zhang2019contextual} don't get. Table \ref{table1} summarizes the different setting of our work and previous models.
\begin{table}[htbp]
	\centering 
	\label{table1} 
	\caption{Comparisons of our setting with previous models.}  
	\begin{tabular}{|c|c|c|c|c|c|}  
		\hline
		setting&context&cascading&position discount&general reward& conservative \\
		\hline
		Conservative UCB\cite{wu2016conservative}&no&no&no&no&yes \\
		CLUCB\cite{kazerouni2016conservative}&yes&no&no&no&yes \\
		CCConUCB\cite{zhang2019contextual}&yes&no&no&yes&no \\
		$C^3$-UCB\cite{li2016contextual}&yes&yes&yes&yes&no \\
		$C^4$-UCB&yes&yes&yes&yes&yes \\
		\hline
	\end{tabular}
\end{table}

\section{Related Work}

Since Auer et al.\cite{auer2002finite} first summarize several fundamental algorithms for the simple bandit and explicitly give the details of the proof which make a great contribution to the bandit field, there have been much recent progress in this field. 

\textbf{Contextual Bandit} Li et al.\cite{li2010contextual} first put the context ($d$-dimension vector) into consideration and put forward the contextual bandit. In this model, the reward of the arm is a linear function of the unknown vector $\theta$. They also show its application in personalization news article recommendation. But there remains regret proof to be completed. Subsequently, Wang et al.\cite{wang2016learning} optimize the contextual bandit to model the hidden features in the bandit and enhance the accuracy in experiments. Abbasi et al.\cite{abbasi2011improved} first give the $\tilde{O}(d\sqrt{T})$ upper bound for the contextual bandit. Gan et al.\cite{gan2019online} study the context the setting given. They investigate what could happen if the context satisfies some probability distribution. And They prove the constant upper bound for it when meeting diverse context constraint.

\textbf{Combinatorial Bandit}   Chen et al.\cite{chen2013combinatorial} first propose the combinatorial bandit, where the agent can choose a group of arms each time and receive a reward corresponding the arm set. A follow-up\cite{chen2016combinatorial} extends the model with a general unlinear reward function such as max and min. They also propose a new discretization technique to bound it. Degnne et al.\cite{degenne2016combinatorial} explore the combinatorial semi-bandit when knowning the covariance of the arm's distribution. Qin et al.\cite{qin2014contextual} first propose the contextual combinatorial bandit and experiments it on the online recommendation. Kveton et al.\cite{kveton2015tight} make use of some special sequences of numbers to further optimize the upper bound for combinatorial semi-bandit.

\textbf{Cascading Bandit}   Kveton et al.\cite{kveton2015cascading} first propose the cascading bandit where you should recommend a list of items to user in order to get a maximum cumulative rewards. And they give a brief $O(\log(n))$ upper bound. Li et al.\cite{li2016contextual} propose the contextual combinatorial cascading bandits ($C^3$-bandit). They incorporate contextual information into the model. They also consider the position in the cascading model and generalize the reward function meeting some common constraints. Gan et al.\cite{gan2020cost} propose a cost-aware cascading bandits model. When agents pull an arm, the agent deserves not only a reward but also a cost. The reward minus cost is the final reward. They solve it by UCB-based method and prove $O(\log(n))$ upper bound.

\textbf{Conservative Bandit}   Wu et al.\cite{wu2016conservative} first propose the conservative bandit. It's a bandit model at each step cumulative rewards are superior than baseline strategy. Kazerouni et al.\cite{kazerouni2016conservative} propose the conservative contextual linear bandit which takes the contextual information into the conservative bandit. Garcelon et al.\cite{garcelon2020improved} utilize the martingale inequality to optimize the confidence set in the conservative-bandit UCB-based algorithms. Also, they simplify and combine the action selection procedure so that the conservative algorithm can do more exploration in the condition of satisfying the conservative constraint. In addition, they also consider the checkpoint situation and design the corresponding algorithm. Zhang et al.\cite{zhang2019contextual} propose the conservative contextual combinatorial bandit, design a algorithm and prove an $O(d^2+d\sqrt{T})$ upper regret bound. They don't make the regret from the conservative mechanism under a constant term.

Beyond that, there are some distinctive creative innovations on bandits. For example, Chen et al.\cite{chen2017interactive} propose the interactive submodular bandit to put the bandit in the interactive situation. Srinivas et al.\cite{srinivas2009gaussian} first utilize the gaussian process regression to model the bandit problem. Soon after, Krause et al.\cite{krause2011contextual} incorporate the context information and design the CGP-UCB algorithm. Lu et al.\cite{lu2020regret} consider the bandit on the causal graph, design the UCB-based and Thompson Sampling-based method to solve it and prove $O(d\sqrt{T})$ upper regret bound. Wang et al.\cite{wang2018regional} study the non-linear reward function of the arm and study the conditions when arms have correlations. Reverdy et al.\cite{reverdy2014modeling} model the human decision-making as the bandit problem. Every arm obeys the gaussian distributions and the agents know the variance. Thus, they design the Upper Credible Limit method to solve this special model. Boursier et al.\cite{boursier2018sic} start to study the multi-agent bandit learning. In this model, there are many agents to choose arms rather than one, and users who choose the same arm can collide and receive no reward.\\

\section{Problem Formulation}

We formulate the problem of \emph{Conservative Contextual Cascading Combinatorial Bandit} as follows. And note that, we follow the notations of \cite{li2016contextual}. Suppose we have a finite set of arm $E=[1,...,L]$ of $L$ \emph{ground items}, also refered to as \emph{base arms}. Let $\Pi^K=\{(a_1,...,a_k):a_1,...a_k \in E, a_i \neq a_j$ for any $i \neq j\}$ be the set of $k$-tuple of distinct items from $E$, also refered to as \emph{super arms}. Each of such tuple is called an action of length $K$. We use $|A|$ to denote the length of action $A$. $S=\Pi^{\leq K} =\cup_{k=1}^K \Pi^K$ means feasible solution with length up to $K$.

Every turn at time $t$, feature vectors $x_{t,a},1 \leq a \leq L$ with $\Vert x_{t,a} \Vert_2 \leq 1$ (also named context) for some base arm are revealed to the learning agent, each feature vector combines the information of user and the base arm. The expected weight $w$ of every base arm is $$w_{t,a}=\theta_*^{\top}x_{t,a}.$$
The weight of every base arm at time $t$ is a random variable:$$w_t(a)=\theta_*^{\top}x_{t,a}+\eta_t,$$
where $\eta$ is a noise which follows the $R$-sub-gaussian distribution. The reward of every super arm $A$ is a function of the weight of base item $w_{t}(a)$, $a \in A$ . Then, the learning agent recommends a list $A_t=(a_1^t,...,a^t_{|A_t|}) \in S$ to the user. The user checks from the top of the list, and stops at the $O_t$-th item under some stopping criterion. Then the learning agent receives rewards, also observes the stopping location $O_t$ and the weight of first $O_t$ base arms $w_t(a_k^t),k \leq O_t$. Hence an item is observed at time $t$ if $a=a_k^t,k \leq O_t$. We don't need the exact form of the reward function just some property.

Also, we introduce the position discount $\gamma_k \in [0,1],1\leq k \leq K$. The every item's final weight has to multiple the $\gamma_k$ to show the difference of location. The item possessing the larger position discount has more influence to the reward function.

When it comes to the conservative mechanism, there is also a set of $|A_0|$ other base arm $A_0=\{L+1,...,L+|A_0|\}$. The cumulative rewards shouldn't be less than a certain fraction of the rewards gained by simply choose the baseline strategy $A_0$, which can be formulated by:
\begin{equation}
\label{1}
\sum_{s=1}^tf(A,w_s) \geq (1-\epsilon)f(A_0,w_s)t,\ \forall t \in [T].
\end{equation}
The expected reward of $A_0$ is $u_0$. The parameter $\epsilon \in [0,1]$ determines the conservative degree. $A_t^*=\arg\max_{A \in \Theta^K}f(A,w_t)$, where $\Theta^K=A_0 \cup A,A \in S$ is the set of feasible solutions.

What's more, in order to make the problem doable, we should make some assumptions which fit the fact.

\begin{assumption}
The reward function of the super arm satisfies following qualities:\\
\textbf{Monotonicity} The expected reward fucntion $f(A,w)$ is non-decreasing with $w$: for any $w,w' \in [0,1]^E$,if $w(a) \leq w'(a)$, we have $f(A,w) \leq f(A,w')$;\\
\textbf{Lipschitz continuity}
The expected reward function $f(A,w)$ is B-Lipschitz continuous with respect to $w$ together with position discount parameters $\gamma_k,k \leq K$. More specifically, for any $w,w' \in [0,1]^E$, we have
$$|f(A,w)-f(A,w')| \leq B \sum_{k=1}^{|A|}\gamma_k|w(a_k)-w'(a_k)|,$$
where $A=(a_1,...,a_{|A|}).$
\end{assumption}
\begin{assumption}
Each element $\eta$  of the noise sequence $\{\eta_t\}_{t=1}^{\infty}$ is conditionally R-sub-gaussian, \emph{i.e.}, $\mathbb{E}[e^{\zeta\eta_t}|a_{1:t},\eta_{1:t-1}]\leq \exp(\zeta^2R^2/2).$
\end{assumption}

\begin{assumption}
$\Vert\theta^* \Vert_2 \leq 1$,$\Vert x_a^t \Vert_2 \leq 1$, and $\langle \theta^*,x_a^t \rangle \in [0,1]$ for all $t$ and all $a \in A_t.$
\end{assumption}

\begin{assumption}
\label{asum4}

There exist $0\leq \Delta_l \leq \Delta_h$ and $0<r_l$ such that, at each round $t$,\\
\centerline{$\Delta_l \leq \alpha f(A_t^*,w_t)-u_0 \leq \Delta_h$, }
where $u_0$ is the expected reward of baseline strategy in every turn.
\end{assumption}

We have an oracle $O_s$ to get the approximate answer every turn which is greater than $\alpha f^*$. Therefore, we use the $\alpha$-regret of action $A$ on time $t$ is 
 $$R^{\alpha}(t,A)=\alpha f^*_t-f(A,w_t),$$
 where $f^*_t=f(A_t^*,w_t)$, $A_t^*=\arg\max_{A \in S}f(A,w_t)$ and $w_t=(\theta_*^\top x_{t,a})_{a \in E}$. Our goal is to minimize the $\alpha$-regret
 $$R^{\alpha}(n)=\mathbb{E}[\sum_{t=1}^nR^{\alpha}(t,A_t)].$$

\section{Algorithms}

We propose two algorithms based on the upper confidence bound strategy to solve the $C^4$-bandit, in both cases when the conservative reward is prescribed and when it's unknown. In the latter situation, we know the baseline strategy's action, but don't know the action's corresponding expected reward.

\subsection{Known Baseline Reward}

\subsubsection{Construction of confidence set}

Because arm's expected weight $w$ is the linear function of $\theta$, so we can use linear least square method to estimate the $\theta$:
$$\hat{\theta} =(X_t^{\top}X_t+\lambda I)^{-1}X_t^{\top}Y_t,$$
where $X_t$ is the matrix of $\{\gamma_k x_{s,a}\},k\in [O_s],s\in[t]$, $Y_t$ is the matrix of $\{\gamma_k w_t(a)\},a\in [O_t],s\in[t]$.

Then, we should control the deviation of this estimation. Therefore, we introduce two lemmas to construct confidence bound of the parameter $\theta$, and calculate the deviation of the upper bound of $w$ (or lower bound) to the its center.
\begin{lemma}
\label{lem1}
Let
$$\beta_t(\delta)=R\sqrt{\ln(\frac{\det(V_t)}{\lambda^d \delta^2})}+\sqrt{\lambda}.$$
Then for any $\delta>0$, with probability at least $1-\delta$, for all $t>0$, we have 
$$\Vert \hat{\theta}-\theta_* \Vert_{V_t} \leq \beta_t(\delta).$$
\end{lemma}
We can construct the following confidence set based on Lemma \ref{lem1}:
$$C_{t+1}=\{\theta \in R^d: \Vert \theta-\hat{\theta} \Vert_{V_t} \leq \beta_{t}(\delta)\}.$$
Note that $\mathbb{P}[\theta \in  C_{t},\forall t \in N] \geq 1-\delta$. And we define the upper confidence bound and lower confidene bound of the expected weight of the arm $a$:
$$U_t(a)=\min\{\hat{\theta}_{t-1}^\top x_{t,a}+\beta_{t-1}(\delta)\Vert \hat{\theta}-\theta_* \Vert_{V_t^{-1}},1\},$$
$$L_t(a)=\max\{\hat{\theta}_{t-1}^\top x_{t,a}-\beta_{t-1}(\delta)\Vert \hat{\theta}-\theta_* \Vert_{V_t^{-1}},0\}.$$
\begin{lemma}
\label{lemma2}
On event $\xi=\{\theta^* \in C_t, \forall t \in N \}$, for any $t \in N$, we have 
$$0 \leq U_t(a)-w_{t,a} \leq 2 \beta_{t-1}(\delta)\Vert \hat{\theta}-\theta_* \Vert_{V_t^{-1}},$$
$$0 \leq w_{t,a}-L_t(a) \leq 2 \beta_{t-1}(\delta)\Vert \hat{\theta}-\theta_* \Vert_{V_t^{-1}}.$$
\end{lemma}
Based on Lemma \ref{lemma2}, we can measure the gap between $U_t(a)$ and $w_{t,a}$, and control the regret. We use the Optimism in the Face of Uncertainty principle, and take the arm with the maximum $U_t$ at each time. Based on this principle, we can control the sub-linear regret in almost every bandit model. $C^4$-bandit is not an exception.

\subsubsection{Guarantee of conservative constraint}
The key question in this section is to guarantee the algorithm's reward is superior to a certain fraction of the baseline strategy with high probability. We consider Ineq.(\ref{1})
$$
\sum_{s=1}^tf(A,w_s) \geq (1-\epsilon)f(A_0,w_s)t,\forall t \in [T].
$$
If the lower bound of LHS is larger than upper bound of RHS, then Ineq.(\ref{1}) holds with high probability. According to Lemma \ref{lemma2}, $w_{t,a}-L_t(a)\geq 0$ with high probability. Using the monotonicity, one can see that $f(A,w_{t,a}) \geq f(A,L_t(a))$. Therefore, we use the following inequality as the conservative constraint:
\begin{equation}
\label{3}
\underbrace{\sum_{s=1}^tf(A,L_t)}_{\text{lower bound of LHS}} \geq \underbrace{(1-\epsilon)u_0t}_{\text{upper bound of RHS}},\ \forall t \in [T].
\end{equation}

Therefore, we use the Ineq.(\ref{3}) to judge if the conservative constraint is satisfied in the algorithm. Each time the algorithm chooses conservative step, it adds $\epsilon u_0$ more to the LHS than it does to the RHS, thus increasing the probability of inequality holds in the subsequent step. And in explorative step, the algorithm utilize the more adding term to do exploration. It's just like that the conservative step is increasing the conservative budget, and the UCB-step is consuming the budget.

Based on above analysis, we now introduce the UCB-based algorithm for the $C^4$-bandit problem. Assume we have the expected reward $u_0$. The algorithm can be separated by three parts: First, update the upper confidence bound and lower confidence bound for every arm. The baseline strategy's upper confidence bound and lower confidence bound are equal to $u_0$. Second, define the best arm $B_t$ using the oracle for the subsequent procedure and update corresponding confidence bound. Although we can't improve the upper bound by using time $t$'s confidence bound, but I think this just can be tighter using some different proving methods. Lastly, judge whether the conservative constraint is satisfied, and do corresponding procedure. If the algorithm satisfies the inequality, then it means the lower confidence bound of $\psi_t$ is larger than the cumulative expected rewards of the baseline strategy. Consequently, we can do the UCB-step not violating the constraint with high probability. If not, we have to do the baseline step to add the conservative budget.

	\begin{algorithm}[H]
		\caption{$C^4$-UCB with Known Conservative Reward.}
		\KwIn{$\gamma_k \in [0,1]_{k \leq K};\delta =\frac{1}{\sqrt{n}}; \lambda \geq C_{\gamma} =\sum_{k=1}^K \gamma_k^2,\epsilon,N_0=D_0=\emptyset,u_0$}
		Initialization:\\
		$\hat{\theta}_0=0,\beta_0(\delta)=1,V_0=\lambda I,X_0=\emptyset,Y_0=\emptyset$\\
		\For{$t =1,2,...,n$}{
		    Obtain  context $x_{t,e}$ for all $e \in E$\\
		    \For{$e \in E$}{		    
		        $U_{t,e}=\min\{ \hat{\theta}_{t-1}^{\top}x_{t,e}+\beta_{t-1}(\delta)\Vert x_{t,e}\Vert_{V_{t-1}^{-1}},1\}$\\
		        $L_{t,e}=\max\{ \hat{\theta}_{t-1}^{\top}x_{t,e}-\beta_{t-1}(\delta)\Vert x_{t,e}\Vert_{V_{t-1}^{-1}},0\}$
		    }
		    $B_t \leftarrow O_s(U_t)$\\
		    \For{$n \in  N_{t-1} \land  e \in A_n$}{
		        $L_{n,e} \leftarrow \max\{0,\hat{\theta}_{t-1}^{\top}x_{n,e}-\beta_{t-1}(\delta)\Vert x_{n,e}\Vert_{V_{t-1}^{-1}}\}$
		    }
		    
		    $L_t \leftarrow \sum_{n \in N_{t-1}} f(A_n,L_n)+f(B_t,L_t)+|D_{t-1}|u_0$
		    
		    \uIf{$L_t \geq (1-\epsilon)tu_0$}{
		        $A_t \leftarrow B_t,N_t \leftarrow N_{t-1} \cup {t}$
		    
		        Play $B_t$ and observe $O_t,w_{t,a^t_k},k \in[O_t]$\\
		    
		        $V_t \leftarrow V_{t-1} +\sum_{k=1}^{O_t} \gamma_k^2x_{t,a^t_k}x_{t,a^t_k}^{\top}$\\
		        $X_t \leftarrow [X_{t-1};\gamma_1x^{\top}_{t,a_1,t};...;\gamma_{O_t}x^{\top}_{t,a_{O_t}^t}]$\\
		        $Y_t \leftarrow [Y_{t-1};\gamma_1\omega_{t,a_1^t};...;\gamma_{O_t}\omega_{t,a_{O_t}^t}]$\\
		        $\hat{\theta} \leftarrow (X_t^{\top}X_t+\lambda I)^{-1}X_t^{\top}Y_t$\\
		        $\beta_t(\delta) \leftarrow R \sqrt{\ln(\det(V_t)/\lambda^d\delta^2)}$
		    }\Else{
		        $A_t \leftarrow A_0,D_t\leftarrow D_{t-1}\cup {t}$
		    }
		}
 	\end{algorithm}

\subsection{Unknown Baseline Reward}
With respect to the unknown baseline reward, we just modify the \textbf{Algorithm 1} to adapt to the new situation. Because we now don't know the expected reward of the baseline strategy, we need to estimate the baseline reward using the confidence set. Then we can get the baseline reward's confidence interval. We introduce the following lemma to solve the conservative constraint.

\begin{lemma}
\label{lemma3}
In unknown baseline reward situation, if
\begin{equation}
\label{eq3}
    \begin{aligned}
        \sum_{n \in N_{t-1}} f(A_n,L_{t})+f(B_t,L_{t})+d_{t-1}f(A_0,U_{t}) \geq (1-\epsilon)tf(A_0,U_{t}).
    \end{aligned}
\end{equation}

holds, then Ineq.(\ref{1}) holds with high probability.
\end{lemma}

\begin{proof}
\begin{itemize}
    \item \textbf{Case 1:} If $d_{t-1}<(1-\epsilon)t$, since $f(A_0,U_{t})>u_0$ with high probability, then Ineq.(\ref{1}) holds with high probability.
    \item \textbf{Case 2:} If  $d_{t-1}\geq(1-\epsilon)t$, then
$\sum_{s=1}^tu_s \geq d_{t-1}u_0 \geq (1-\epsilon)u_0t.$ Ineq.(\ref{1}) must hold.
\end{itemize}
\end{proof}

The Lemma \ref{lemma3} gives the judging criteria whether satisfying the conservative constraint with high probability in the unknown baseline reward situation. Thus, we modify the conservative constraint in \textbf{Algorithm 1} using the Ineq.(\ref{eq3}). The new algorithm is given in \textbf{Algorirhm 2} in Appendix.

\section{Regret Analysis}

Several definitions will be given before introducing the main theorem. Let $p_{t,A}$ denote the probability of full observation of $A$, denote $p^*=\min_{t\in[1,T],A\in S} p_{t,A}$. Let $d$ represent the dimension of the context. $K$ is the maximum volume of super arm. $B$ is the Lipschitz coefficient. The $\eta$ obeys the $R$-sub-gaussian distribution. $n_T$ and $d_T$ are denoted as the number of UCB-step and conservative step respectively. 
\subsection{Regret in known baseline reward situation}

\begin{theorem}\label{thm1}
If $\lambda \geq L$, then the following regret bound of \textbf{Algorithm 1} is satisfied with probability at least $1-\delta$:
\begin{equation}
    \begin{aligned}
    R^{\alpha}(T)
    &\leq \frac{2\sqrt{2}B}{p^*}(R\sqrt{\ln[(1+C_\gamma T/(\lambda d))^dT]}+\sqrt{\lambda})\\
&\quad \quad\sqrt{TKd\ln(1+C_\gamma T/(\lambda d))}+\alpha \sqrt{T}+(\frac{\Omega}{\epsilon u_0}+1)\Delta_h\\
    &=O(d\sqrt{TK}\ln(C_\gamma T)+(\frac{\Omega}{\epsilon u_0}+1)\Delta_h),
    \end{aligned}
\end{equation}
where $\Omega$ is a constant depending on the problem which has the value of
$$\Omega = \frac{ 442368B^4R^4K^2d^4(1+C_{\gamma}/(\lambda d))^{1/2} }{(p^*)^4(\epsilon u_0+\Delta_l)^3}  +(1-\epsilon)u_0.$$
\end{theorem}

\begin{proof} (sketch) After running \textbf{Algorithm 1} for $T$ rounds, the cumulative regret can be bounded as follows,
\begin{equation}
\begin{aligned}
R^{\alpha}(T)&=\mathbb{E}[\sum_{t=1}^T[\alpha f(A_t^*,w_t)-f(A_t,w_t)]]\\
    &=\mathbb{E}[\sum_{t\in N_T}[\alpha f(A_t^*,w_t)-f(A_t,w_t)]+\sum_{t\in D_T}[ \alpha f(A_t^*,w_t)-u_0]]\\
    &\leq \mathbb{E}[\sum_{t\in N_T}[f(A_t,U_t)-f(A_t,w_t)]+\sum_{t\in D_T}[f(A_t^*,w_t)-u_0]]\\
    &\leq \mathbb{E}[\sum_{t\in N_T}[f(A_t,U_t)-f(A_t,w_t)]]+d_T\Delta_h.
\end{aligned}
\end{equation}

The former part can be bounded by Lemma \ref{lem4}, the latter part can be bounded by Theorem \ref{thm2}.
\end{proof}

\begin{remark}
From the Theorem \ref{thm1}, the first term in the regret bound is the regret of $C^3$-UCB, which grows at rate of $O(\sqrt{T}\log(T))$, the second term indicates the loss due to the conservative constraint in the algorithm which only grows in finite number of rounds. It means that the conservative mechanism in $C^4$-bandit doesn't change the order of the regret. Furthermore, the regret bound indicates that the smaller $\epsilon$, the larger regret. This greatly matches our intuition that the agent who is more conservative has to suffer more regret because of smaller $\epsilon$. If the baseline strategy has lower expected reward, it also has to suffer more regret.
\end{remark}

\begin{lemma}\label{lem4} The cumulative regret of optimistic time step until time step $T$ satisfies the  following with probability at least $1-\delta$:
\begin{equation}
\begin{aligned}
\mathbb{E}[\sum_{i=1}^{n_T}[f(A_t,U_t)-f(A_t,w_t)]]&\leq \frac{2\sqrt{2}B}{p^*}(R\sqrt{\ln[(1+C_\gamma T/(\lambda d))^dT]}+\sqrt{\lambda})\\
&\quad \quad\sqrt{TKd\ln(1+C_\gamma T/(\lambda d))}+\alpha \sqrt{T}.\\
\end{aligned}
\end{equation}
\end{lemma}
\begin{proof}
Because the confidence set is changing only in the optimistic exploration step. Just use the conclusion from \cite{li2016contextual}, as $n_T \leq T$, and replace the $n_T$ with $T$ to make an upper bound.
\end{proof}
 Before we prove Theorem \ref{thm2}, we bring in some important lemmas to help the proof procedure.
\begin{lemma}
\label{lem5}
For any $t\geq 1$,
$$f(A_t,U_t)-f(A_t,L_t)\leq 4B \sum_{k=1}^{|A_t|}\gamma_k\beta_{t-1}(\delta)\Vert x_{t,a_k^t}\Vert_{V^{-1}_{t-1}}.$$
\end{lemma}
\begin{proof}(sketch) It can be proved based on the B-lipschitz continuity of $f$ and Lemma \ref{lemma2}.
\end{proof}
\begin{lemma}
\label{lem6}
For any $t\geq1$, if $\lambda \geq C_\gamma$,then
$$\sum_{s=1}^t \sum_{k=1}^{O_t}\Vert \gamma_k x_{s,a_k^s} \Vert_{V_{s-1}^{-1}}^2 \leq 2d\ln(1+C_{\gamma}t/(\lambda d)).$$
\end{lemma}
\begin{lemma}
\label{lem7}
$\det(V_t)$ is increasing with respect to t and
$\det(V_t) \leq (\lambda+C_{\gamma}t/d)^d$
\end{lemma}
\begin{remark}
Lemma \ref{lem6} and Lemma \ref{lem7} are conclusions from \cite{li2016contextual}.
\end{remark}

\begin{theorem}\label{thm2}
On event $\xi=\{\theta^* \in c_t,\ \forall t \in N\}$, for any $T\geq 1$, we have
\begin{equation}
    \begin{aligned}
d_T &\leq   \frac{ 442368B^4R^4K^2d^4(1+C_{\gamma}/(\lambda d))^{1/2} }{(p^*)^4(\epsilon u_0+\Delta_l)^3 \epsilon u_0}  +(1/\epsilon-1)+1.
    \end{aligned}
\end{equation}
\end{theorem}
\begin{proof}(sketch) Suppose $t$ is the last round the conservative policy is played before time $T$, then 
$d_T=d_{t-1}+1.$ Then according to \textbf{Algorithm 1}, it is satisfied that 
$$\sum_{n \in N_{t-1}} f(A_n,L_t)+f(B_t,L_t)+|D_{t-1}|u_0 <(1-\epsilon)u_0t.$$
Note that t can be denoted as $t=n_{t-1}+d_{t-1}+1$. By dropping $f(B_t,L_t)$, and rearranging the terms, we have
\begin{equation}
    \begin{aligned}
    &\epsilon d_{t-1} u_0\\
    &<[1-(1+n_{t-1})\epsilon]u_0+n_{t-1}u_0-\sum_{n \in N_{t-1}}f(A_n,L_t)\\
    &=[1-(1+n_{t-1})\epsilon]u_0+\sum_{n \in N_{t-1}}[u_0-f(A_n,U_n)+f(A_n,U_n)-f(A_n,U_t)\\
    &\quad \quad +f(A_n,U_t)-f(A_n,L_t)]\\
    &\leq [1-(1+n_{t-1})\epsilon]u_0+\sum_{n \in N_{t-1}}[u_0-f(A_n,U_n)+f(A_n,U_n)-f(A_n,L_n)\\
    &\quad \quad +f(A_n,U_t)-f(A_n,L_t)]\\
    &\leq [1-(1+n_{t-1})\epsilon]u_0 - n_{t-1} \Delta_l+ \frac{8B}{p^*}\beta_{t-1}(\delta)\sum_{n\in N_{t-1}}\sum_{k=1}^{|O_t|}\Vert \gamma_kx_{t,a} \Vert_{V_{t-1}^{-1}}\\
    &\leq [1-(1+n_{t-1})\epsilon]u_0 - n_{t-1} \Delta_l+ \frac{8B}{p^*}\beta_{t-1}(\delta)\sqrt{(\sum_{t=1}^{n_{t-1}}O_t)(\sum_{t=1}^{n_{t-1}}\sum_{k=1}^{O_t}\Vert\gamma_k x_{t,a} \Vert^2_{V_{t-1}^{-1}})}\\
    &\leq [1-(1+n_{t-1})\epsilon]u_0 - n_{t-1} \Delta_l+\frac{8\sqrt{2}B}{p^*}(R\sqrt{ln[(1+C_{\gamma}n_{t-1}/(\lambda d))^dn_{t-1}]}+\sqrt{\lambda})\\
    &\quad \sqrt{n_{t-1}Kd\ln(1+C_{\gamma}n_{t-1}/(\lambda d))}.
    \end{aligned}\\
\end{equation}
The second inequality is from the Lemma \ref{lem5}. The third inequality is from definition of $p^*$. The fourth inequality is by the mean inequality. The last inequality is from the definition of $\beta$, Lemma \ref{lem6} and Lemma \ref{lem7}.

Then, the RHS of the inequality is a function of $n_{t-1}$ which has the maximum denoted as $\Omega$ and
$$\Omega=   \frac{ 442368B^4R^4K^2d^4(1+C_{\gamma}/(\lambda d))^{1/2} }{(p^*)^4(\epsilon u_0+\Delta_l)^3}  +(1-\epsilon)u_0.$$
Finally, we can get an upper bound of $d_T$.

\end{proof}

\subsection{Regret in the unknown baseline reward situation}
\begin{theorem}
\label{thm3}
If $\lambda \geq L$, then the following regret bound of \textbf{Algorithm 2} is satisfied with probability at least $1-\delta$:
\begin{equation}
    \begin{aligned}
    R^{\alpha}(T)
    &\leq \frac{2\sqrt{2}B}{p^*}(R\sqrt{\ln[(1+C_\gamma T/(\lambda d))^dT]}+\sqrt{\lambda})\\
&\quad \quad\sqrt{TKd\ln(1+C_\gamma T/(\lambda d))}+\alpha \sqrt{T}+(\frac{\Omega}{\epsilon u_0}+1)\Delta_h\\
    &=O(d\sqrt{TK}\ln(C_\gamma T)+(\frac{\Omega}{\epsilon u_0}+1)\Delta_h).
    \end{aligned}
\end{equation}
where $\Omega$ is a constant depending on the problem which has the value of
\begin{equation}
    \begin{aligned}
    \Omega&=
    \max \{\frac{442368B^4R^4d^4K^2(1+C_{\gamma}/(\lambda d))^{1/2}}{(p^*)^4 \epsilon^3(u_0+BK\gamma_1)^3}+(1-\epsilon)(u_0+BK \gamma_1),\\
    &\frac{442368B^4R^4d^4K^2(1+C_{\gamma}/(\lambda d))^{1/2}}{(p^*)^4 \epsilon^3}+(1-\epsilon)\}.
    \end{aligned}
\end{equation}
\end{theorem}
\begin{remark}
The Theorem \ref{thm3} indicates that in unknown reward situation, the regret is the same as known reward situation: it can be decomposed into an upper bound from the optimistic step and a constant term from the conservative step. The only difference between two situations is the constant term is different. It indicates that the knowledge of $u_0$ doesn't change the order of regret in the cascading bandit model. However, they both don't grow with the time horizon $T$. The proof follows the same procedures as Theorem \ref{thm2}. However, it needs a little extra technique skills. What's more, in the procedure of proof, we also get a tight bound for the conservative contextual combinatorial bandit. The complete proof is attached in the Appendix. 
\end{remark}

\subsection{Improved bound for the conservative contextual combinatorial bandit}
\begin{theorem}\label{thm1}
If $\lambda \geq L$, then the following regret bound of conservative contextual combinatorial bandit\cite{zhang2019contextual} is satisfied with probability at least $1-\delta$:
\begin{equation}
    \begin{aligned}
    R^{\alpha}(T)\leq O(d\sqrt{T}\log(\frac{TK}{d}))+O(1)
    \end{aligned}
\end{equation}
\end{theorem}
Zhang et al. didn't bound the regret from the conservative mechanism in the unknown baseline reward situation under a constant term. This is because the core step in the proof,
$$f(A_0,U_t) \leq f(A_0,w)+2B\beta_{t-1}\sqrt{\sum_{e \in A_0} \Vert x_{0,e}\Vert_{V_{t-1}^{-1}}^2},$$
where $C_{t-1}$ is a term growing with time $t$. We use the technique from our proof of regret, according to their definetion of Lipschitz continuity, we improve this bound by:
\begin{equation}
    \begin{aligned}
    f(A_0,U_t)- f(A_0,w)&\leq B\Vert U_t-w \Vert_2 \\
    &= B\sqrt{\sum_{k \in A_0} (U_t(k)-w(k))^2}\\
    &\leq B\sqrt{K}
    \end{aligned}
\end{equation}
The first inequality is because Lipschitz continuity, the second is as $w \in [0,1]^E$. We don't translate the $U-w$ to $\Vert x \Vert_{V_{t-1}^{-1}}$ using the confidence bound. However, we use the range of $w$ to bound it. This is because we usually use the sum of $\Vert x \Vert_{V_{t-1}^{-1}}$ from $1$ to $t$ to bound the overall regret, and this sum of $t$ over $V_{t-1}$ can provide a sublinear regret. But this time, we don't possess the sum from $1$ to $t$, so if we bound it to a constant, it can't provide a linear regret depending on $T$. Instead, if we transform it to the $\Vert x \Vert_{V_{t-1}^{-1}}$, we again bring in the dependence on $\beta_t$, which is related with $T$. Its coupling with $n_{t-1}$ (the number of explorative step) can lead to complicated from.  That's what we don't want to see. However, if we bound the $f(A_0,U)$ to a constant, we can use a function which has a unique maximum to bound the regret from the conservative mechanism to a constant term.
\section{Experiments}

In this section, we evaluate our algorithms on synthetic data. We compare our algorithm with $C^3$-UCB which doesn't have the conservative mechanism. Also, we compare our algorithm with different parameter to validate how these parameters influence our algorithm.

\subsection{Setting}
We use the disjunctive objective for experiments, which means the user stops at the first attractive item position. The reward function has the pattern of 
$f(A,w)=\sum_{K=1}^{|A|}\gamma_k \prod_{i=1}^{k-1}(1-w(a_i))w(a_k).$
This function satisfies the monotonicity and Lipschitz continuity. The proof can be viewed in \cite{li2016contextual}. We randomly choose $\theta$ with $\Vert \theta \Vert_2=1$ (choose every dimension from $N(0,1)$ and then normalize entire vector) and let $\theta_*=(\theta/2,1/2)$ as the optimal $\theta$ at the begin of the game. And at each time step,  we randomly generate $x_{t,a}$ with $\Vert x'_{t,a} \Vert_2=1$, let $x_{t,a}=(x'_{t,a},1)$ be the contextual information. Then we can make $\theta_*x_{t,a} \in [0,1]$. The context has 20 dimensions. At each turn, we generate 200 contexts, and the feasible solution's largest number is 4 items. When the agent pulls the arm, the environment utilize the $\theta$ and the corresponding $x_{t,a}$ to calculate the action's expected  weight. Then sample every weight with bernoulli distribution, stop at the first attactive item and give feedback to the agent. Since in our setting, the value of reward function tends to be very high, so we let $u_0$ equal to a relatively high value to make baseline strategy is comparable with other random action. As for the $C^4$-UCB without known baseline reward, we sample its value from $N(u_0,0.1)$ every turn. And we use UCB algorithm for the simple bandit to calculate its upper confidence bound. Other default hyper-parameters are as follows: $\delta=0.1$, $\lambda=0.1$, $u_0=0.7$, $T=4*10^4$, $\gamma_k=1$ (no position discount).

\subsection{Results}

\begin{figure}
\centering
\begin{minipage}{.46\textwidth}  
  \centering
  \includegraphics[width=1\linewidth]{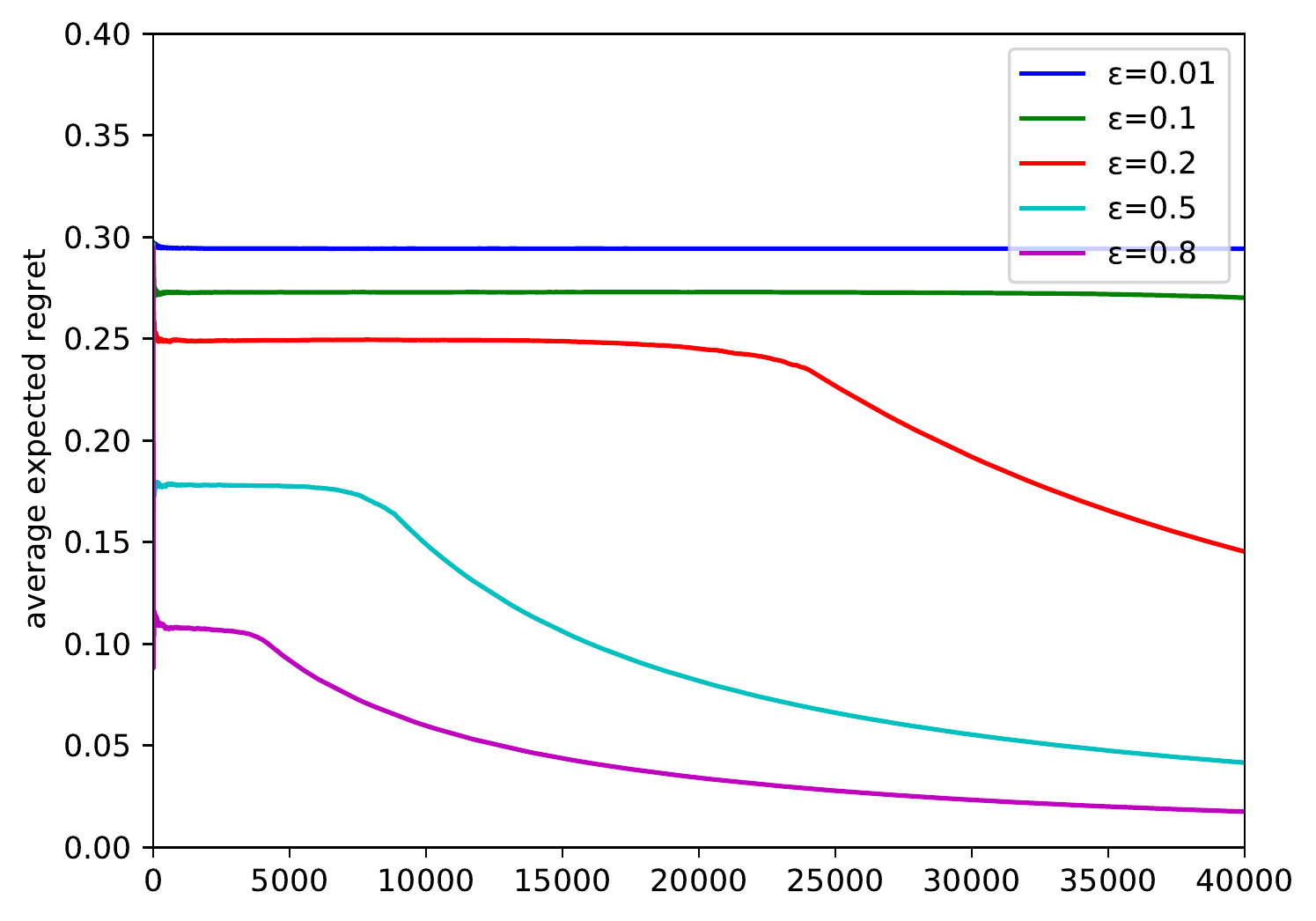}
  \caption{Average expected regret of $C^4$-UCB with different $\epsilon$.}
  \label{fig1}
\end{minipage}%
\ \ \ 
\begin{minipage}{.46\textwidth}
  \centering
  \includegraphics[width=1\linewidth]{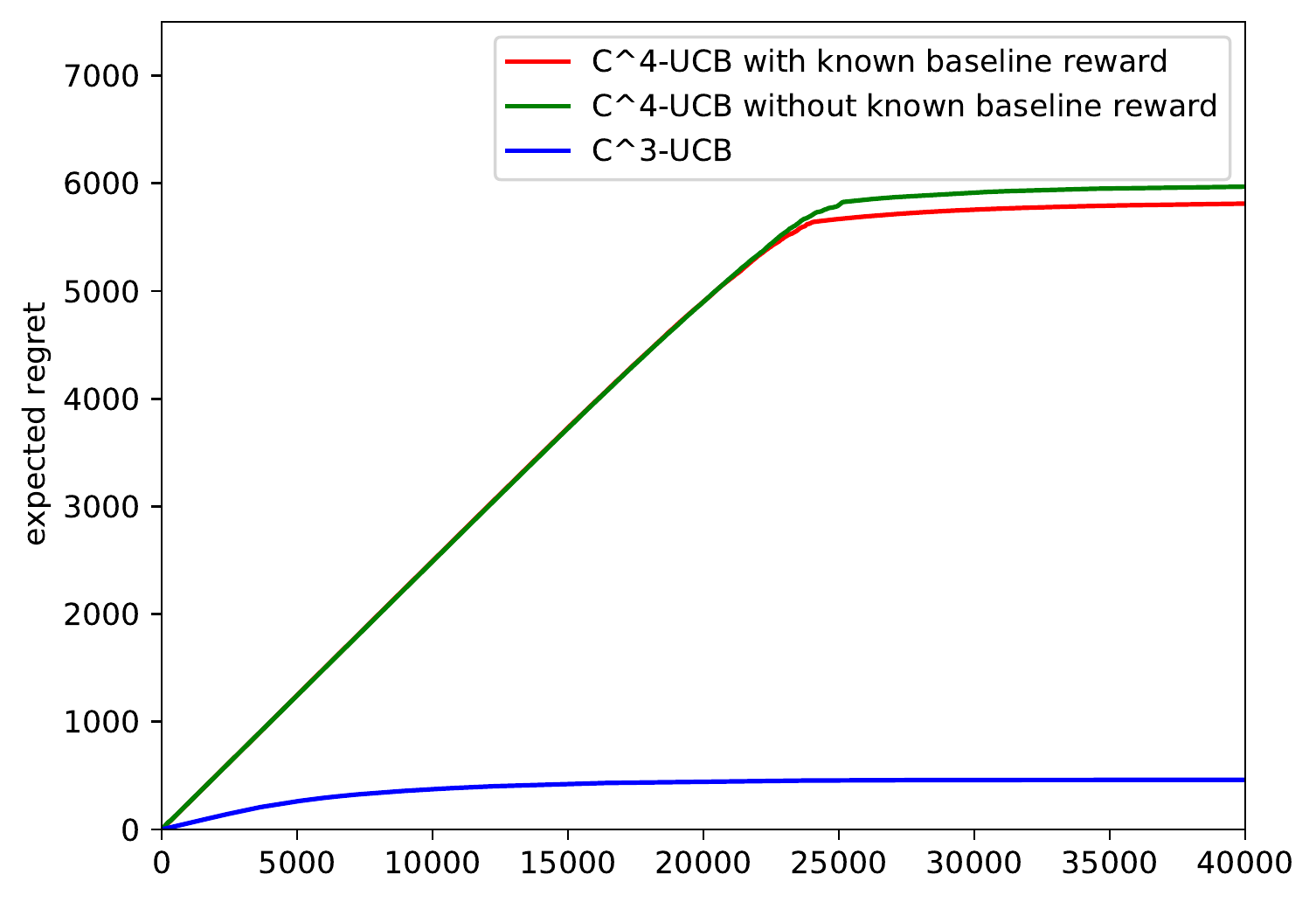}
  \caption{Cumulative expected regret of distinct models.}
  \label{fig2}
\end{minipage}
\end{figure}

In the first experiment, we compare our algorithm $C^4$-UCB to $C^3$-UCB which doesn't satisfy the conservative constraint by cumulative regret. Just as Fig.\ref{fig2} shows, due to the extra constraint for the conservative mechanism, the $C^4$-UCB bears the larger regret over time. However, it still has the sub-linear regret, which means the problem is still learnable. This validate our regret analysis that the $C^4$-UCB has the $\tilde{O}(\sqrt{T})$ regret. The $C^4$-UCB without known baseline reward suffers more regret than algorithm with known reward because of the inaccurate estimation of $u_0$. In the first 20000 rounds, the two $C^4$-UCB algorithms almost coincide. This indicates that in early rounds, the $A_0$'s confidence set's shrinking make little influence on the regret. We check data from the initial phase. And we discover that lower confidence bounds of the normal arms are too small to satisfy the conservative constraint for the known baseline reward situation, not to mention unknown. They both store budget merely from the conservative step, and consume them in the explorative step. What's more, we can also get from the picture the three algorithm all have the same trend of increasing, which confirms that they share the same order of regret. It fits our analysis of regret.

In the second experiment, we evaluate how $\epsilon$ influences the algorithm's performance. Fig.\ref{fig1} shows that the UCB algorithm which has the larger $\epsilon$ bear smaller regret. This is because larger $\epsilon$ means agents can endure more errors. The agents have the more comfortable constraint. This leads to a smaller regret. Also, the average regret all declines but the time they start to decline is different. The algorithm which has smaller $\epsilon$ start to decline lately. Especially, the algorithms whose $\epsilon=0.01$ and $0.1$ even don't decline for the first $4*10^4$ rounds. Table \ref{table2} explains the cause of this situation. The Table \ref{table2} shows the exact number of the conservative step and optimistic step for the $C^4$-UCB as $\epsilon$ varies. It shows that the algorithm which has the larger $\epsilon$ plays more explorative step. The algorithm which has the smaller $\epsilon$ don't want to suffer a little risks of safety, and have to do less explorative steps. This leads to the confidence set shrinking slowly. Therefore, their probability of breaking the conservative constraint isn't small enough, which leads to this situation.

\begin{minipage}{\textwidth}
 \begin{minipage}[t]{0.46\textwidth}
  \centering
     \makeatletter\def\@captype{table}\makeatother\caption{The contrast of the number of ucb step and conservative step when $\epsilon$ varies.}
       \begin{tabular}{cccc}
\toprule
$\epsilon$ & ucb step & conservative step  \\
\midrule
0.01    & 400  &39600 \\
0.1 & 4539& 35641\\
0.2    & 21995& 18005 \\
0.5    & 35999& 4001     \\
0.8     & 39252& 748\\
\bottomrule
	\end{tabular}
	\label{table2}
  \end{minipage}
  \begin{minipage}[t]{0.46\textwidth}
   \centering
        \makeatletter\def\@captype{table}\makeatother\caption{The contrast of the number of ucb step and conservative step when $u_0$ varies.}
         \begin{tabular}{cccc}        
\toprule
$u_0$ & ucb step & conservative step  \\
\midrule
0.2    & 27653  &12347 \\
0.5 & 23483& 16517\\
0.7    & 22288& 17712 \\
0.9    & 19979& 20021     \\
0.95     & 19539& 20461\\
\bottomrule
	  \end{tabular}
	  \label{table3}
   \end{minipage}
\end{minipage}


\begin{figure}
\centering
\begin{minipage}{.46\textwidth}  
  \centering
  \includegraphics[width=1\linewidth]{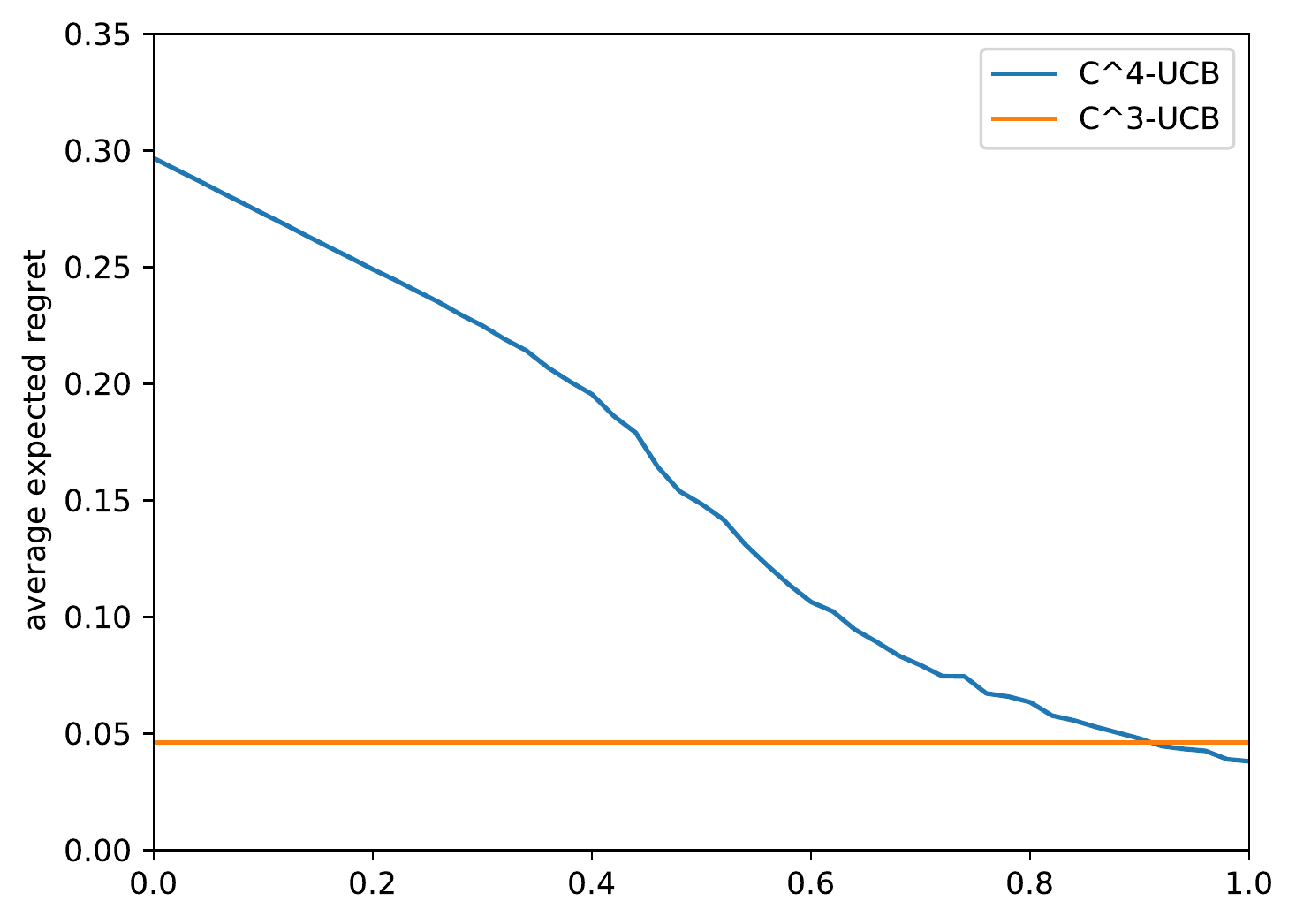}
  \caption{Average regret for varying $\epsilon$ and $T = 10^4$ and $\delta = 0.01$.}
  \label{fig3}
\end{minipage}%
\ \ \ 
\begin{minipage}{.46\textwidth}
  \centering
  \includegraphics[width=1\linewidth]{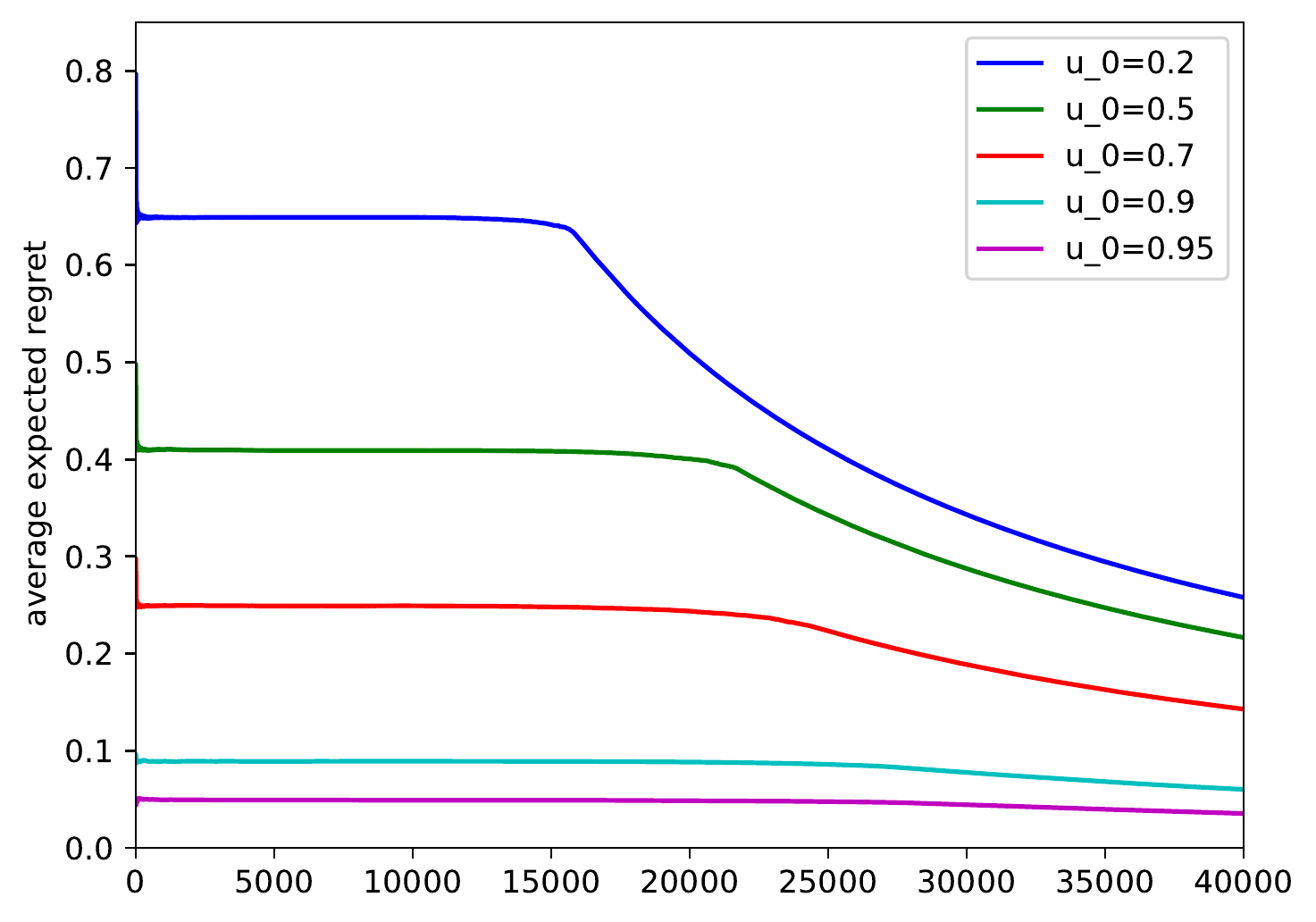}
  \caption{Average expected regret of $C^4$-UCB with different $u_0$.}
  \label{fig4}
\end{minipage}
\end{figure}

In the third experiment,  we compare the average expected regret of our algorithm to $C^3$-UCB for the first $10^4$ rounds as $\epsilon$ varies. We sample $\epsilon$ at 0.02 intervals. In Fig.\ref{fig3}, the $C^3$-UCB's regret doesn't change as $\epsilon$ varies, because it doesn't take the conservative constraint into the consideration. However, the $C^4$-UCB declines as $\epsilon$ becomes larger. Since larger $\epsilon$ means larger tolerance.

In the fourth experiment, we investigate how $u_0$ influences our algorithm's performance. We may have the following question: if $u_0$ becomes larger, the baseline reward becomes higher leading to lower regret. However, this also reduces the probability of satisfying conservative constraint, so it reduces the explorative step and makes confidence set shrink slower. And it finally leads to larger regret in this perspective. This opposite feedback to regret makes us difficult to analyze it. This experiment helps us to figure out this situation. Fig.\ref{fig4} shows that the algorithm which has larger $u_0$ suffers smaller regret. And Table \ref{table3} shows that the algorithm which has larger $u_0$ plays less explorative step. Combining both results, one can see that although the algorithm with larger $u_0$ plays less exploration, it still bears smaller regret. This is because $u_0$'s direct impact on regret has a bigger influence on the regret than by indirectly reducing the number of explorative step.

\section{Discussion and Conclusion}
In this paper, we introduce a new novel setting of bandits, referred to as conservative contextual combinatorial cascading bandit ($C^4$-bandit). And we design the corresponding $C^4$-UCB to solve it. We study how the conservative mechanism influences the design of algorithm in the cascading bandit and how it changes the regret term. We demonstrate that the conservative mechanism doesn't change the order of the regret of the UCB-based method in cascading bandits, just adding a new constant term. By the way, we bound the regret from conservative mechanism in the contextual combinatorial bandit under a constant term for the first time. In future, there will be work such as in non-stationary environment and the proof of lower bound to be completed.

%
%
%
\bibliographystyle{splncs04}
\bibliography{mybibliography}

\newpage
\appendix

\section{Algorithm 2}
	\begin{algorithm}[H]
		\caption{$C^4$-UCB with Unknown Conservative Reward.}
		\KwIn{$\gamma_k \in [0,1]_{k \leq K};\delta =\frac{1}{\sqrt{n}}; \lambda \geq C_{\gamma} =\sum_{k=1}^K \gamma_k^2,\epsilon,N_0=D_0=\emptyset$}
		Initialization:\\
		$\hat{\theta}_0=0,\beta_0(\delta)=1,V_0=\lambda I,X_0=\emptyset,Y_0=\emptyset$\\
		\For{$t =1,2,...,n$}{
		    Obtain  context $x_{t,e}$ for all $e \in E\cup A_0$\\
		    
		    \For{$e \in E \cup A_0$}{		    
		        $U_{t,e}=\min\{ \hat{\theta}_{t-1}^{\top}x_{t,e}+\beta_{t-1}(\delta)\Vert x_{t,e}\Vert_{V_{t-1}^{-1}},1\}$\\
		        $L_{t,e}=\max\{ \hat{\theta}_{t-1}^{\top}x_{t,e}-\beta_{t-1}(\delta)\Vert x_{t,e}\Vert_{V_{t-1}^{-1}},0\}$
		    }

		      $B_t=(a_1^t,...,a^t_{|A_t|}) \leftarrow O_s(U_t)$

		    \For{$n \in  N_{t-1} \land  e \in A_n$}{
		        $L_{n,e} \leftarrow \max\{0,\hat{\theta}_{t-1}^Tx_{n,e}-\beta_{t-1}(\delta)\Vert x_{n,e}\Vert_{V_{t-1}^{-1}}\}$
		    }
		    
		    $L_t \leftarrow \sum_{n \in N_{t-1}} f(A_n,L_n)+f(B_t,L_t)+|D_{t-1}|f(A_0,U_{t,0})$
		    
		    \uIf{$L_t \geq (1-\alpha)tf(A_0,U_{t})$}{
		        $A_t \leftarrow B_t,N_t \leftarrow N_{t-1} \cup {t}$
		    
		        Play $B_t$ and observe $O_t,w_{t,a^t_k},k \in[O_t]$\\
		    
		        $V_t \leftarrow V_{t-1} +\sum_{k=1}^{O_t} \gamma_k^2x_{t,a^t_k}x_{t,a^t_k}^{\top}$\\
		        $X_t \leftarrow [X_{t-1};\gamma_1x^{\top}_{t,a_1,t};...;\gamma_{O_t}x^{\top}_{t,a_{O_t}^t}]$\\
		        $Y_t \leftarrow [Y_{t-1};\gamma_1\omega_{t,a_1^t};...;\gamma_{O_t}\omega_{t,a_{O_t}^t}]$\\
		        $\hat{\theta} \leftarrow (X_t^{\top}X_t+\lambda I)^{-1}X_t^{\top}Y_t$\\
		        $\beta_t(\delta) \leftarrow R \sqrt{\ln(\det(V_t)/\lambda^d\delta^2)}$
		    }\Else{
		        $A_t \leftarrow A_0,D_t\leftarrow D_{t-1}\cup {t}$
		    }
		}
	\end{algorithm}

\section{Proof of Theorem 2}

\begin{proof}

Suppose $t$ is the last round the conservative policy is played, then 
$$d_T=d_{t-1}+1.$$
Then according to \textbf{Algorithm 1}, it is satisfied that 
$$\sum_{n \in N_{t-1}} f(A_n,L_n)+f(B_t,L_t)+|D_{t-1}|u_0 <(1-\epsilon)u_0t.$$ 
Note that $t$ can be denoted as $t=n_{t-1}+d_{t-1}+1$. By dropping $f(B_t,L_t)$, and rearranging the terms, we have

\begin{equation}
    \begin{aligned}
    &\epsilon d_{t-1} u_0\\
    &<[1-(1+n_{t-1})\epsilon]u_0+n_{t-1}u_0-\sum_{n \in N_{t-1}}f(A_n,L_t)\\
    &=[1-(1+n_{t-1})\epsilon]u_0+\sum_{n \in N_{t-1}}[u_0-f(A_n,U_n)+f(A_n,U_n)-f(A_n,U_t)\\
    &\quad \quad +f(A_n,U_t)-f(A_n,L_t)]\\
    &\leq [1-(1+n_{t-1})\epsilon]u_0+\sum_{n \in N_{t-1}}[u_0-f(A_n,U_n)+f(A_n,U_n)-f(A_n,L_n)\\
    &\quad \quad +f(A_n,U_t)-f(A_n,L_t)]\\
    &\leq [1-(1+n_{t-1})\epsilon]u_0 - n_{t-1} \Delta_l+ \frac{8B}{p^*}\beta_{t-1}(\delta)\sum_{n\in N_{t-1}}\sum_{k=1}^{|O_t|}\Vert \gamma_kx_{t,a} \Vert_{V_{n-1}^{-1}}\\
    &\leq [1-(1+n_{t-1})\epsilon]u_0 - n_{t-1} \Delta_l+ \frac{8B}{p^*}\beta_{t-1}(\delta)\sqrt{(\sum_{t=1}^{n_{t-1}}O_t)(\sum_{t=1}^{n_{t-1}}\sum_{k=1}^{O_t}\Vert\gamma_k x_{t,a} \Vert^2_{V_{t-1}^{-1}})}\\
    &\leq [1-(1+n_{t-1})\epsilon]u_0 - n_{t-1} \Delta_l+\frac{8\sqrt{2}B}{p^*}(R\sqrt{ln[(1+C_{\gamma}n_{t-1}/(\lambda d))^dn_{t-1}]}+\sqrt{\lambda})\\
    &\quad \sqrt{n_{t-1}Kd\ln(1+C_{\gamma}n_{t-1}/(\lambda d))}.
    \end{aligned}\\
\end{equation}

    When $n_{t-1} \geq \frac{\lambda d (e^{\lambda/(R^2d)}-1)}{C_{\gamma}}$,
\begin{equation}
    \begin{aligned}
    &\epsilon d_{t-1} u_0\\
    &\leq [1-(1+n_{t-1})\epsilon]u_0 - n_{t-1} \Delta_l+\frac{16\sqrt{2}B}{p^*}R\sqrt{\ln[(1+C_{\gamma}n_{t-1}/(\lambda d))^dn_{t-1}]}\\
    &\quad \sqrt{n_{t-1}Kd\ln(1+C_{\gamma}n_{t-1}/(\lambda d))}.
    \end{aligned}\\
\end{equation}

Because the rightmost is a function of $n_{t-1}$ with a pattern of $$f(x)=-Cx+A\sqrt{\ln[(1+Ex)^dx]}\sqrt{x\ln(1+Ex)}+D,$$
where $A=16BR\sqrt{2Kd}/p^*$, $C=(\epsilon u_0+\Delta_l)$, $D=(1-\epsilon)u_0$, $E=C_{\gamma}/(\lambda d).$\\
We assume $n_{t-1} \geq 1$, then we analyze $f(x)$,
\begin{equation}
    \begin{aligned}
    f(x)&=-Cx+A\sqrt{\ln[(1+Ex)^dx]}\sqrt{x\ln(1+Ex)}+D\\
    &\leq -Cx+A\sqrt{d\ln[(1+Ex)x]}\sqrt{x\ln(1+Ex)}+D\\
    &= -Cx+A\sqrt{d}\sqrt{x}\ln[(1+Ex)x]+D\\
    &\leq -Cx+2A\sqrt{d}\sqrt{x}\ln[\sqrt{1+E}x]+D.
    \end{aligned}\\
\end{equation}

Because $x^{1/4} \geq \ln x$, when $x\geq 6000$, thus when $x\geq \frac{6000}{\sqrt{1+E}}$ \emph{i.e.}, $x \geq \frac{6000}{\sqrt{1+C_{\gamma}/(\lambda d)}}$,

$$f(x) \leq -Cx+2A\sqrt{d}(1+E)^{1/8}x^{3/4}+D.$$

Let $$g(x)=-Cx+2A\sqrt{d}(1+E)^{1/8}x^{3/4}+D,$$
$$g'(x)=-C+\frac{3A\sqrt{d}(1+E)^{1/8}}{2x^{1/4}},$$
$$g''(x)=-\frac{15A\sqrt{d}(1+E)^{1/8}}{8x^{5/4}}<0.$$
Therefore, $g'(x)$ is a monotonic decreasing function with $x$, and $g(x)$ first increases and then decreases with $x$.\\
Let $$g'(x_0)=0.$$
$$x_0=(\frac{3A\sqrt{d}(1+E)^{1/8}}{2C})^4.$$
$g(x_0)$ is the maximum value of $g(x)$. Substitute $x_0$ in the $g(x)$, one can see that the maximum value of $g(x)$ is
\begin{equation}
    \begin{aligned}
g(x) &\leq -C(\frac{3A\sqrt{d}(1+E)^{1/8}}{2C})^4+2A\sqrt{d}(1+E)^{1/8}(\frac{3A\sqrt{d}(1+E)^{1/8}}{2C})^3+D\\
&=-\frac{81A^4 d^2(1+E)^{1/2}}{16C^3} +\frac{54A^4d^2(1+E)^{1/2}}{8C^3}
 +D\\
&=\frac{27A^4d^2(1+E)^{1/2}}{16C^3}+D.
    \end{aligned}
\end{equation}
Therefore, we then substitute the $A,B,C,D$ into it and get the maximum  $\Omega$ of $f(x)$.\\
\begin{equation}
    \begin{aligned}
f(x) &\leq   \frac{ 442368^4R^4K^2d^4(1+C_{\gamma}/(\lambda d))^{1/2} }{(p^*)^4(\epsilon u_0+\Delta_l)^3}  +(1-\epsilon)u_0.
    \end{aligned}
\end{equation}
Then,
$$\epsilon d_{t-1}u_0 <\Omega,$$
$$d_{t-1} \leq \frac{\Omega}{\epsilon u_{0}},$$
$$d_T=d_{t-1}+1 <\frac{\Omega}{\epsilon u_0}+1=\Omega'.$$
Proof is completed.
\end{proof}

\section{Proof of Theorem 3}

\begin{proof}

 If we choose the conservative arm in time step $t$, then according to \textbf{Algorithm 2}, it is satisfied that

\begin{equation}
    \begin{aligned}
    \sum_{n \in N_{t-1}} &f(A_n,L_{t})+f(B_t,L_{t})+d_{t-1}f(A_0,U_{t})<(1-\epsilon)tf(A_0,U_{t}).
    \end{aligned}
\end{equation}

Suppose $t$ is the last time the algorithm plays the conservative step, then
$$d_{t-1}=d_T-1,$$
and 
\begin{equation}
    \begin{aligned}
    \sum_{n \in N_{t-1}} &f(A_n,L_{t})+f(B_t,L_{t})+(d_{t-1}-(1-\epsilon)t)f(A_0,U_{t})<0.
    \end{aligned}
\end{equation}
Then we get
\begin{equation}
    \begin{aligned}
    &\epsilon d_{t-1} f(A_0,U_{t})\\
    &<[1-(1+n_{t-1})\epsilon]f(A_0,U_{t})+n_{t-1}f(A_0,U_{t})-\sum_{n \in N_{t-1}}f(A_n,L_t)\\
    & \leq [1-(1+n_{t-1})\epsilon]f(A_0,U_{t})+\sum_{n \in N_{t-1}}[f(A_0,U_{t})-f(A_n,U_{n})\\
    &\quad \quad +f(A_n,U_{n})-f(A_n,U_{t})+f(A_n,U_{t})-f(A_n,L_{t})]\\
    &\leq [1-(1+n_{t-1})\epsilon]f(A_0,U_{t})+\sum_{n \in N_{t-1}}[f(A_n,U_{n})-f(A_n,L_n)+B\sum_{k=1}^{|A_n|}\gamma_k|U_{t,e}-L_{t,e}|]\\
    &= [1-(1+n_{t-1})\epsilon]f(A_0,U_{t})+\sum_{n \in N_{t-1}}[8B\beta_{t-1}(\delta)\sum_{k=1}^{|A_n|}\Vert\gamma_k x_{t,a} \Vert_{V_{n-1}^{-1}}]\\
    &\leq [1-(1+n_{t-1})\epsilon]f(A_0,U_{t}) + \frac{8B}{p^*}\beta_{t-1}(\delta)\sum_{n\in N_{t-1}}\sum_{k=1}^{|O_n|}\Vert \gamma_kx_{t,a} \Vert_{V_{n-1}^{-1}}\\
    &\leq [1-(1+n_{t-1})\epsilon]f(A_0,U_{t}) + \frac{8B}{p^*}\beta_{t-1}(\delta)\sqrt{(\sum_{t=1}^{n_{t-1}}O_t)(\sum_{t=1}^{n_{t-1}}\sum_{k=1}^{O_t}\Vert\gamma_k x_{t,a} \Vert^2_{V_{t-1}^{-1}})}\\
    &\leq [1-(1+n_{t-1})\epsilon]f(A_0,U_{t}) +\frac{8\sqrt{2}B}{p^*}(R\sqrt{\ln[(1+C_{\gamma}n_{t-1}/(\lambda d))^dn_{t-1}]}+\sqrt{\lambda})\\
    &\quad \sqrt{n_{t-1}Kd\ln(1+C_{\gamma}n_{t-1}/(\lambda d))}.
    \end{aligned}\\
\end{equation}
We divide the problem into two cases:\\
\textbf{Case 1:} If $[1-(1+n_{t-1})\epsilon]<0$, then we get
\begin{equation}
    \begin{aligned}
    &\epsilon d_{t-1} u_0\leq [1-(1+n_{t-1})\epsilon]u_0  +\frac{16\sqrt{2}B}{p^*}R\sqrt{\ln[(1+C_{\gamma}n_{t-1}/(\lambda d))^dn_{t-1}]}\\
    &\quad \sqrt{n_{t-1}Kd\ln(1+C_{\gamma}n_{t-1}/(\lambda d))}.
    \end{aligned}\\
\end{equation}
Therefore, use the same technology as the known $u_0$ situation just removing the $-\Delta_l n_{t-1}$,
\begin{equation}
    \begin{aligned}
\epsilon d_{t-1} u_0 &\leq   \frac{ 442368B^4R^4K^2d^4(1+C_{\gamma}/(\lambda d))^{1/2} }{(p^*)^4(\epsilon u_0)^3}  +(1-\epsilon)u_0.
    \end{aligned}
\end{equation}
\textbf{Case 2:} If $[1-(1+n_{t-1})\epsilon]\geq 0$, because 
\begin{equation}
\label{e21}
    \begin{aligned}
    f(A_0,U_t)-f(A_0,w) &\leq B\sum_{k=1}^{|A_0|}\gamma_k |U_t(k)-w(k)| \leq BK \gamma_1\\
    f(A_0,U_t) &\leq u_0+BK \gamma_1\\
    \end{aligned}
\end{equation}

when $n_{t-1} \geq \frac{\lambda d (e^{\lambda/(R^2d)}-1)}{C_{\gamma}}$, we get
\begin{equation}
    \begin{aligned}
    & \epsilon d_{t-1}u_0\\
     &\leq [1-(1+n_{t-1})\epsilon](u_0+BK\gamma_1) +\frac{8\sqrt{2}B}{p^*}(R\sqrt{\ln[(1+C_{\gamma}n_{t-1}/(\lambda d))^dn_{t-1}]}\\
 &\quad \quad +\sqrt{\lambda})\sqrt{n_{t-1}Kd\ln(1+C_{\gamma}n_{t-1}/(\lambda d))}\\
 &\leq [1-(1+n_{t-1})\epsilon](u_0+BK\gamma_1) +\frac{16\sqrt{2}B}{p^*}(R\sqrt{\ln[(1+C_{\gamma}n_{t-1}/(\lambda d))^dn_{t-1}]})\\
 &\quad \quad \sqrt{n_{t-1}Kd\ln(1+C_{\gamma}n_{t-1}/(\lambda d))}\\
    \end{aligned}
\end{equation}

The rightmost of the inequality is a function of pattern
$$f(x)=-Cx+A\sqrt{\ln[(1+Ex)^dx]}\sqrt{x\ln(1+Ex)}+D,$$
where $A=16BR\sqrt{2Kd}/p^*$, $C=\epsilon u_0+ \epsilon BK \gamma_1$, $D=(1-\epsilon)(u_0+BK\gamma_1)$,
 $E=C_{\gamma}/(\lambda d).$\\
Therefore, we use the technical skill from the proof of Theorem \ref{thm2},
$$f(x)\leq \frac{27A^4d^2(1+E)^{1/2}}{16C^3}+D.$$

\begin{equation}
    \begin{aligned}
    f(x)_{\max}&=\frac{442368B^4R^4d^4K^2(1+C_{\gamma}/(\lambda d))^{1/2}}{(p^*)^4 \epsilon^3(u_0+BK\gamma_1)^3}+(1-\epsilon)(u_0+BK \gamma_1).
    \end{aligned}
\end{equation}

\begin{equation}
    \begin{aligned}
    \epsilon d_{t-1}u_0 &\leq \frac{442368B^4R^4d^4K^2(1+C_{\gamma}/(\lambda d))^{1/2}}{(p^*)^4 \epsilon^3(u_0+BK\gamma_1)^3}+(1-\epsilon)(u_0+BK \gamma_1).
    \end{aligned}
\end{equation}

From all of the above,we get

\begin{equation}
\label{eq26}
    \begin{aligned}
        \epsilon d_{t-1}u_0 &\leq \max \{\frac{442368B^4R^4d^4K^2(1+C_{\gamma}/(\lambda d))^{1/2}}{(p^*)^4 \epsilon^3(u_0+BK\gamma_1)^3}+(1-\epsilon)(u_0+BK \gamma_1),\\
    &\frac{442368B^4R^4d^4K^2(1+C_{\gamma}/(\lambda d))^{1/2}}{(p^*)^4 \epsilon^3}+(1-\epsilon)\}.
    \end{aligned}
\end{equation}
Let $\Omega$ denote RHS of Ineq.(\ref{eq26}), using Lemma \ref{lem4},
\begin{equation}
    \begin{aligned}
    R^{\alpha}(T)
    &\leq \frac{2\sqrt{2}B}{p^*}(R\sqrt{\ln[(1+C_\gamma T/(\lambda d))^d T]+\sqrt{\lambda}})\\
    &\quad \quad\sqrt{TKd\ln(1+C_\gamma T/(\lambda d)}+(\frac{\Omega}{\epsilon u_0}+1)\Delta_{h}.
    \end{aligned}
\end{equation}
Proof is completed.\\
What's more, by the Ineq.(\ref{e21}), we can bound the regret from the conservative constraint in conservative contextual combinatorial bandit under a constant term\cite{zhang2019contextual}.
\end{proof}

\end{document}


\title{Conservative Contextual Combinatorial Cascading Bandit}
%
%
\author{Kun Wang\inst{1} \and
Canzhe Zhao\inst{2} }

\authorrunning{Kun Wang and Canzhe Zhao}

%
\institute{Shanghai Jiaotong University, Shanghai, China
\and
Shandong University, School of Software, Shandong, China\\
\email{wangkun8512@sjtu.edu.cn}\\
\email{zcz@mail.sdu.edu.cn}
}

\section{Algorithm 2}
	\begin{algorithm}[H]
		\caption{$C^4$-UCB with Unknown Conservative Reward}
		\KwIn{$\gamma_k \in [0,1]_{k \leq K};\delta =\frac{1}{\sqrt{n}}; \lambda \geq C_{\gamma} =\sum_{k=1}^K \gamma_k^2,\epsilon,N_0=D_0=\emptyset$}
		Initialization:\\
		$\hat{\theta}_0=0,\beta_0(\delta)=1,V_0=\lambda I,X_0=\emptyset,Y_0=\emptyset$\\
		\For{$t =1,2,...,n$}{
		    Obtain  context $x_{t,e}$ for all $e \in E\cup A_0$\\
		    
		    \For{$e \in E \cup A_0$}{		    
		        $U_{t,e}=\min\{ \hat{\theta}_{t-1}^{\top}x_{t,e}+\beta_{t-1}(\delta)\Vert x_{t,e}\Vert_{V_{t-1}^{-1}},1\}$\\
		        $L_{t,e}=\max\{ \hat{\theta}_{t-1}^{\top}x_{t,e}-\beta_{t-1}(\delta)\Vert x_{t,e}\Vert_{V_{t-1}^{-1}},0\}$
		    }

		    \uIf{$f(A_0,U_{t,t})> \max_{A \in  \Theta^K  \backslash A_0}f(A,U_{t,t})$}{
		        $B_t=A_0$}\Else{
		        $B_t=(a_1^t,...,a^t_{|A_t|}) \leftarrow O_s(U_t)$
		    }
		    
		    \For{$n \in  N_{t-1} \land  e \in A_n$}{
		        $L_{n,e} \leftarrow \max\{0,\hat{\theta}_{t-1}^Tx_{n,e}-\beta_{t-1}(\delta)\Vert x_{n,e}\Vert_{V_{t-1}^{-1}}\}$
		    }
		    
		    $\psi_t \leftarrow \sum_{n \in N_{t-1}} f(A_n,L_n)+f(B_t,L_t)+|D_{t-1}|f(A_0,U_{t,0})$
		    
		    \uIf{$\psi_t \geq (1-\alpha)tf(A_0,U_{t,0})$}{
		        $A_t \leftarrow B_t,N_t \leftarrow N_{t-1} \cup {t}$
		    
		        Play $B_t$ and observe $O_t,w_{t,a^t_k},k \in[O_t]$\\
		    
		        $V_t \leftarrow V_{t-1} +\sum_{k=1}^{O_t} \gamma_k^2x_{t,a^t_k}x_{t,a^t_k}^{\top}$\\
		        $X_t \leftarrow [X_{t-1};\gamma_1x^{\top}_{t,a_1,t};...;\gamma_{O_t}x^{\top}_{t,a_{O_t}^t}]$\\
		        $Y_t \leftarrow [Y_{t-1};\gamma_1\omega_{t,a_1^t};...;\gamma_{O_t}\omega_{t,a_{O_t}^t}]$\\
		        $\hat{\theta} \leftarrow (X_t^{\top}X_t+\lambda I)^{-1}X_t^{\top}Y_t$\\
		        $\beta_t(\delta) \leftarrow R \sqrt{\ln(\det(V_t)/\lambda^d\delta^2)}$
		    }\Else{
		        $A_t \leftarrow A_0,D_t\leftarrow D_{t-1}\cup {t}$
		    }
		}
	\end{algorithm}

\section{Proof of Theorem 2}

\begin{theorem}\label{thm2}
On event $\xi=\{\theta^* \in c_t,\forall t \in N\}$,for any $T\geq 1$,we have
\begin{equation}
    \begin{aligned}
d_T \leq -\frac{ 27 BR\sqrt{2kd}d^2(1+C_{\gamma}/(\lambda d))^{1/2} }{4p^*(\epsilon u_0+\Delta_l)^3\epsilon u_0}  +(1/\epsilon-1)u_0+1.
    \end{aligned}
\end{equation}

\end{theorem}

\begin{proof}

Suppose $t$ is the last round the conservative policy is played, then 
$$d_T=d_{t-1}+1.$$
Then according to \textbf{Algorithm 1}, it is satisfied that 
$$\sum_{n \in N_{t-1}} f(A_n,L_n)+f(B_t,L_t)+|D_{t-1}|u_0 <(1-\epsilon)u_0t.$$ 
Note that $t$ can be denoted as $t=n_{t-1}+d_{t-1}+1$. By dropping $f(B_t,L_t)$, and rearranging the terms,we have

\begin{equation}
    \begin{aligned}
    &\epsilon d_{t-1} u_0\\
    &<[1-(1+n_{t-1})\epsilon]u_0+n_{t-1}u_0-\sum_{n \in N_{t-1}}f(A_n,L_n)\\
    &=[1-(1+n_{t-1})\epsilon]u_0+\sum_{n \in N_{t-1}}[u_0-f(A_n,U_t)+f(A_n,U_t)-f(A_n,L_t)]\\
    &\leq [1-(1+n_{t-1})\epsilon]u_0+\sum_{n \in N_{t-1}}[u_0-f(A_n,U_t)+B\sum_{k=1}^{|A_n|}\gamma_k|U_{n,e}-L_{n,e}|]\\
    &= [1-(1+n_{t-1})\epsilon]u_0+\sum_{n \in N_{t-1}}[u_0-f(A_n,U_t)+4B\beta_{t-1}(\delta)\sum_{k=1}^{|A_n|}\Vert\gamma_k x_{t,a} \Vert_{V_{t-1}^{-1}}]\\
    &\leq [1-(1+n_{t-1})\epsilon]u_0 - n_{t-1} \Delta_l+ \frac{4B}{p^*}\beta_{t-1}(\delta)\sum_{n\in N_{t-1}}\sum_{k=1}^{|O_t|}\Vert \gamma_kx_{t,a} \Vert_{V_{t-1}^{-1}}]\\
    &\leq [1-(1+n_{t-1})\epsilon]u_0 - n_{t-1} \Delta_l+ \frac{4B}{p^*}\beta_{t-1}(\delta)\sqrt{(\sum_{t=1}^{n_{t-1}}O_t)(\sum_{t=1}^{n_{t-1}}\sum_{k=1}^{O_t}\Vert\gamma_k x_{t,a} \Vert^2_{V_{t-1}^{-1}})}\\
    &\leq [1-(1+n_{t-1})\epsilon]u_0 - n_{t-1} \Delta_l+\frac{4\sqrt{2}B}{p^*}(R\sqrt{\ln[(1+C_{\gamma}n_{t-1}/(\lambda d))^dn_{t-1}]}+\sqrt{\lambda})\\
    &\quad \sqrt{n_{t-1}Kd\ln(1+C_{\gamma}n_{t-1}/(\lambda d))}.
    \end{aligned}\\
\end{equation}

When $n_{t-1} \geq \frac{e^{\lambda/(R^2 d^2)}}{(1+C_{\gamma}/(\lambda d))^{1/d}}$,

\begin{equation}
    \begin{aligned}
    &\epsilon d_{t-1} u_0\\
    &\leq [1-(1+n_{t-1})\epsilon]u_0 - n_{t-1} \Delta_l+\frac{8\sqrt{2}B}{p^*}R\sqrt{\ln[(1+C_{\gamma}n_{t-1}/(\lambda d))^dn_{t-1}]}\\
    &\quad \sqrt{n_{t-1}Kd\ln(1+C_{\gamma}n_{t-1}/(\lambda d))}.
    \end{aligned}\\
\end{equation}

Because the rightmost is a function of $n_{t-1}$ with a pattern of $$f(x)=-Cx+A\sqrt{\ln[(1+Ex)^dx]}\sqrt{x\ln(1+Ex)}+D,$$
where $A=8BR\sqrt{2Kd}/p^*$,$C=(\epsilon u_0+\Delta_l)$,$D=(1-\epsilon)u_0$,$E=C_{\gamma}/(\lambda d).$\\
Then we analyze the $f(x)$,
\begin{equation}
    \begin{aligned}
    f(x)&=-Cx+A\sqrt{\ln[(1+Ex)^dx]}\sqrt{x\ln(1+Ex)}+D\\
    &\leq -Cx+A\sqrt{d\ln[(1+Ex)x]}\sqrt{x\ln(1+Ex)}+D\\
    &\leq -Cx+A\sqrt{d}\sqrt{x}\ln[(1+Ex)x]+D\\
    &\leq -Cx+2A\sqrt{d}\sqrt{x}\ln[\sqrt{1+E}x]+D.
    \end{aligned}\\
\end{equation}

Because $x^{1/4} \geq \ln(x)$, when $x\geq 6000$, thus when $x\geq \frac{6000}{\sqrt{1+E}}$ whicn means $x \geq \frac{6000}{\sqrt{1+C_{\gamma}/(\lambda d)}}$,

$$f(x) \leq -Cx+2A\sqrt{d}(1+E)^{1/8}x^{3/4}+D.$$

Let $$g(x)=-Cx+2A\sqrt{d}(1+E)^{1/8}x^{3/4}+D,$$
$$g'(x)=-C+\frac{3A\sqrt{d}(1+E)^{1/8}}{2x^{1/4}},$$
$$g''(x)=-\frac{15A\sqrt{d}(1+E)^{1/8}}{8x^{5/4}}<0.$$
Therefore, $g'(x)$ is a monotone decreasing function with $x$, and $g(x)$ first increase then decrease with $x$.\\
Let $$g'(x_0)=0.$$
$$x_0=(\frac{3A\sqrt{d}(1+E)^{1/8}}{2C})^4.$$
$g(x_0)$ is the maximum value of $g(x)$. Substitute $x_0$ in the $g(x)$,we get the maximum value of $g(x)$ is
\begin{equation}
    \begin{aligned}
g(x) &\leq -C(\frac{3A\sqrt{d}(1+E)^{1/8}}{2C})^4+2A\sqrt{d}(1+E)^{1/8}(\frac{3A\sqrt{d}(1+E)^{1/8}}{2C})^3+D\\
&=-\frac{81A^4 d^2(1+E)^{1/2}}{16C^3} +\frac{27A^4d^2(1+E)^{1/2}}{8C^3}
 +D\\
&=-\frac{27A^4d^2(1+E)^{1/2}}{16C^3}+D.
    \end{aligned}
\end{equation}
Therefore, we then substitute the $A,B,C,D$ into it, we get the maximum  $\Omega$ of $f(x)$.\\
\begin{equation}
    \begin{aligned}
f(x) &\leq   -\frac{ \frac{108 BR\sqrt{2Kd}}{p^*}d^2(1+C_{\gamma}/(\lambda d))^{1/2} }{(\epsilon u_0+\Delta_l)^3}  +(1-\epsilon)u_0.
    \end{aligned}
\end{equation}
Then,
$$\epsilon d_{t-1}u_0 <\Omega,$$
$$d_{t-1} \leq \frac{\Omega}{\epsilon u_{0}},$$
$$d_T=d_{t-1}+1 <\frac{\Omega}{\epsilon u_0}+1=\Omega'.$$
Therefore,
\begin{equation}
    \begin{aligned}
    R^{\alpha}(T)
&\leq \frac{2\sqrt{2}B}{p^*}(R\sqrt{\ln[(1+C_\gamma N_T/(\lambda d))^dN_T]}+\sqrt{\lambda})\\
&\quad \quad\sqrt{N_TKd\ln(1+C_\gamma N_T/(\lambda d))}+\alpha \sqrt{N_T}\\
    &\quad+(\frac{\Omega}{\epsilon u_0}+1)\Delta_{h}.
    \end{aligned}
\end{equation}
\end{proof}

\section{Proof of Theorem 3}

\begin{theorem}
If $\lambda \geq L$, then the following regret bound of \textbf{Algorithm 2} is satisfied with probability at least $1-\delta$
\begin{equation}
    \begin{aligned}
    R^{\alpha}(T)
    &\leq \frac{2\sqrt{2}B}{p^*}(R\sqrt{\ln[(1+C_\gamma T/(\lambda d))^dT]}+\sqrt{\lambda})\\
&\quad \quad\sqrt{TKd\ln(1+C_\gamma T/(\lambda d))}+\alpha \sqrt{T}+(\frac{\Omega}{\epsilon u_0}+1)\Delta_h\\
    &=O(d\sqrt{TK}\ln(C_\gamma T)+(\frac{\Omega}{\epsilon u_0}+1)\Delta_h),
    \end{aligned}
\end{equation}
where $\Omega$ is a constant depending on the problem which have the value of
\begin{equation}
    \begin{aligned}
    \Omega&=
    \max\{-\frac{27(8BR\sqrt{2Kd}/p^*)^4d^2(1+C_{\gamma}/(\lambda d))^{1/2}}{16(\epsilon u_0+\Delta_l+2\epsilon B\beta_{t-1}(\delta)\gamma_1 K/\sqrt{\lambda})^3}
    +(1-\epsilon)(u_0+\frac{2B\beta_{t-1}(\delta)\gamma_1 K}{\sqrt{\lambda}}),\\
    &\quad \quad \quad-\frac{ 27 BR\sqrt{2kd}d^2(1+C_{\gamma}/(\lambda d))^{1/2} }{4p^*(\epsilon u_0+\Delta_l)^3}  +(1-\epsilon)u_0\}.
    \end{aligned}
\end{equation}
\end{theorem}

\begin{proof}

 If we choose the conservative arm in time step $t$, then according to \textbf{Algorithm 2}, it is satisfied that

\begin{equation}
    \begin{aligned}
    \sum_{n \in N_{t-1}} &f(A_n,L_{t,n})+f(B_t,L_{t,t})+d_{t-1}f(A_0,U_{s,0})<(1-\epsilon)tf(A_0,U_{s,0}).
    \end{aligned}
\end{equation}

Suppose $t$ is the last time the algorithm plays the conservative step, then
$$d_{t-1}=d_T-1,$$
and 
\begin{equation}
    \begin{aligned}
    \sum_{n \in N_{t-1}} &f(A_n,L_{t})+f(B_t,L_{t})+(d_{t-1}-(1-\epsilon)t)f(A_0,U_{t})<0.
    \end{aligned}
\end{equation}
Then we get
\begin{equation}
    \begin{aligned}
    &\epsilon d_{t-1} f(A_0,U_{t})\\
    &<[1-(1+n_{t-1})\epsilon]f(A_0,U_{t})+n_{t-1}f(A_0,U_{t})-\sum_{n \in N_{t-1}}f(A_n,L_n)\\
    &=[1-(1+n_{t-1})\epsilon]f(A_0,U_{t})+\sum_{n \in N_{t-1}}[f(A_0,U_{t})-f(A_n,U_{t})+f(A_n,U_{t})-f(A_n,L_{t})]\\
    &\leq [1-(1+n_{t-1})\epsilon]f(A_0,U_{t})+\sum_{n \in N_{t-1}}[f(A_0,U_{t})-f(A_n,U_{t})+B\sum_{k=1}^{|A_n|}\gamma_k|U_{n,e}-L_{n,e}|]\\
            \end{aligned}
\end{equation}
\begin{equation}
    \begin{aligned}
    &= [1-(1+n_{t-1})\epsilon]f(A_0,U_{t})+\sum_{n \in N_{t-1}}[f(A_0,U_{t})-f(A_n,U_{t})\\
    &\quad \quad +4B\beta_{t-1}(\delta)\sum_{k=1}^{|A_n|}\Vert\gamma_k x_{t,a} \Vert_{V_{t-1}^{-1}}]\\
    &\leq [1-(1+n_{t-1})\epsilon]f(A_0,U_{t}) - n_{t-1} \Delta_l+ \frac{4B}{p^*}\beta_{t-1}(\delta)\sum_{n\in N_{t-1}}\sum_{k=1}^{|O_t|}\Vert \gamma_kx_{t,a} \Vert_{V_{t-1}^{-1}}]\\
    &\leq [1-(1+n_{t-1})\epsilon]f(A_0,U_{t}) - n_{t-1} \Delta_l+ \frac{4B}{p^*}\beta_{t-1}(\delta)\sqrt{(\sum_{t=1}^{n_{t-1}}O_t)(\sum_{t=1}^{n_{t-1}}\sum_{k=1}^{O_t}\Vert\gamma_k x_{t,a} \Vert^2_{V_{t-1}^{-1}})}\\
    &\leq [1-(1+n_{t-1})\epsilon]f(A_0,U_{t}) - n_{t-1} \Delta_l+\frac{4\sqrt{2}B}{p^*}(R\sqrt{\ln[(1+C_{\gamma}n_{t-1}/(\lambda d))^dn_{t-1}]}+\sqrt{\lambda})\\
    &\quad \sqrt{n_{t-1}Kd\ln(1+C_{\gamma}n_{t-1}/(\lambda d))}.
    \end{aligned}\\
\end{equation}
The third inequality is because$f(A_0,U_{t})-f(A_n,U_{t}) \leq f(A_0,U_{t})- \alpha f(A^*,w^*) \leq u_0 - \alpha f(A^*,w^*) \leq -\Delta_l$.\\
Because
\begin{equation}
    \begin{aligned}
    f(A_0,U)-f(A_0,w) &\leq B\sum_{k=1}^{|A_0|} \gamma_k |U-w|\\
    &\leq2B \beta_{t-1}(\delta)\sum_{k=1}^{|A_0|}\gamma_k \Vert x_{t,a} \Vert_{V^{-1}_{t-1}}\\
    & \leq \frac{2B\beta_{t-1}(\delta)\gamma_1 K}{\sqrt{\lambda}}.
    \end{aligned}
\end{equation}

$$f(A_0,U) \leq u_0+\frac{2B\beta_{t-1}(\delta)\gamma_1 K}{\sqrt{\lambda}}.$$

If $[1-(1+n_{t-1})\epsilon]<0$, then we get
\begin{equation}
    \begin{aligned}
    &\epsilon d_{t-1} u_0\\
    &\leq [1-(1+n_{t-1})\epsilon]u_0 - n_{t-1} \Delta_l +\frac{8\sqrt{2}B}{p^*}R\sqrt{\ln[(1+C_{\gamma}n_{t-1}/(\lambda d))^dn_{t-1}]}\\
    &\quad \sqrt{n_{t-1}Kd\ln(1+C_{\gamma}n_{t-1}/(\lambda d))}.
    \end{aligned}\\
\end{equation}
Therefore, we get the same conclusion as the known $u_0$,
\begin{equation}
    \begin{aligned}
d_{t-1} &\leq   -\frac{ \frac{108 BR\sqrt{2Kd}}{p^*}d^2(1+C_{\gamma}/(\lambda d))^{1/2} }{(\epsilon u_0+\Delta_l)^3}  +(1-\epsilon)u_0.
    \end{aligned}
\end{equation}
If $[1-(1+n_{t-1})\epsilon]\geq 0$, When $n_{t-1} \geq \frac{e^{\lambda/(R^2 d^2)}}{(1+C_{\gamma}/(\lambda d))^{1/d}}$,we get
\begin{equation}
    \begin{aligned}
    & \epsilon d_{t-1}u_0\\ &<(1-\epsilon)(u_0+\frac{2B\beta_{t-1}(\delta)\gamma_1 K}{\sqrt{\lambda}})-n_{t-1}(\epsilon u_0+\Delta_l+\frac{2\epsilon B\beta_{t-1}(\delta)\gamma_1 K}{\sqrt{\lambda}})\\
    &+\frac{8\sqrt{2}B}{p^*}R\sqrt{\ln[(1+C_{\gamma}n_{t-1}/(\lambda d))^dn_{t-1}]}\sqrt{n_{t-1}Kd\ln(1+C_{\gamma}n_{t-1}/(\lambda d))}.
    \end{aligned}
\end{equation}

The rightmost of the inequality is a function of pattern
$$f(x)=-Cx+A\sqrt{\ln[(1+Ex)^dx]}\sqrt{x\ln(1+Ex)}+D,$$
where $A=8BR\sqrt{2Kd}/p^*$,$C=\epsilon u_0+\Delta_l+\frac{2\epsilon B\beta_{t-1}(\delta)\gamma_1 K}{\sqrt{\lambda}}$,$D=(1-\epsilon)(u_0+\frac{2B\beta_{t-1}(\delta)\gamma_1 K}{\sqrt{\lambda}})$,
$E=C_{\gamma}/(\lambda d).$\\
Therefore, we use the conclusion of Theorem 3,
$$f(x)\leq -\frac{27A^4d^2(1+E)^{1/2}}{16C^3}+D.$$

\begin{equation}
    \begin{aligned}
    f(x)_{max}&=-\frac{27(8BR\sqrt{2Kd}/p^*)^4d^2(1+C_{\gamma}/(\lambda d))^{1/2}}{16(\epsilon u_0+\Delta_l+\frac{2\epsilon B\beta_{t-1}(\delta)\gamma_1 K}{\sqrt{\lambda}})^3}+(1-\epsilon)(u_0+\frac{2B\beta_{t-1}(\delta)\gamma_1 K}{\sqrt{\lambda}}).
    \end{aligned}
\end{equation}

\begin{equation}
    \begin{aligned}
    \epsilon d_{t-1}u_0 &\leq -\frac{27(8BR\sqrt{2Kd}/p^*)^4d^2(1+C_{\gamma}/(\lambda d))^{1/2}}{16(\epsilon u_0+\Delta_l+\frac{2\epsilon B\beta_{t-1}(\delta)\gamma_1 K}{\sqrt{\lambda}})^3}+(1-\epsilon)(u_0+\frac{2B\beta_{t-1}(\delta)\gamma_1 K}{\sqrt{\lambda}}).
    \end{aligned}
\end{equation}

From all of the above,we get

\begin{equation}
    \begin{aligned}
        \epsilon d_{t-1} u_0&\leq \max\{-\frac{27(8BR\sqrt{2Kd}/p^*)^4d^2(1+C_{\gamma}/(\lambda d))^{1/2}}{16(\epsilon u_0+\Delta_l+\frac{2\epsilon B\beta_{t-1}(\delta)\gamma_1 K}{\sqrt{\lambda}})^3}\\
    &+(1-\epsilon)(u_0+\frac{2B\beta_{t-1}(\delta)\gamma_1 K}{\sqrt{\lambda}}),\\
    &-\frac{ \frac{108 BR\sqrt{2Kd}}{p^*}d^2(1+C_{\gamma}/(\lambda d))^{1/2} }{(\epsilon u_0+\Delta_l)^3}  +(1-\epsilon)u_0\}.
    \end{aligned}
\end{equation}
Let $\Omega$ denote this maximum, 
\begin{equation}
    \begin{aligned}
    R^{\alpha}(T)
    &\leq \frac{2\sqrt{2}B}{p^*}(R\sqrt{\ln[(1+C_\gamma N_T/(\lambda d)^d)N_T]+\sqrt{\lambda}})\\
    &\quad \quad\sqrt{N_TKd\ln(1+C_\gamma N_T/(\lambda)}+(\frac{\Omega}{\epsilon u_0}+1)\Delta_{h}.
    \end{aligned}
\end{equation}
Proof is completed.
\end{proof}